\documentclass[twoside,11pt]{article}
 
\usepackage[abbrvbib, preprint]{jmlr2e}
\usepackage{graphicx}
\usepackage{amssymb}
\usepackage{color}
\usepackage[normalem]{ulem}
\usepackage{moreverb,url}

\renewcommand*{\cite}[1]{\citep{#1}}

\usepackage{lastpage}
\jmlrheading{21}{2020}{1-\pageref{LastPage}}{9/19; Revised 8/20}{10/20}{19-804}{Oliver Kroemer, Scott Niekum, and George Konidaris}

\ShortHeadings{A Review of Robot Learning for Manipulation}{Kroemer, Niekum, and Konidaris}
\firstpageno{1}

\newcommand\blfootnote[1]{%
  \begingroup
  \renewcommand\thefootnote{}\footnote{#1}%
  \addtocounter{footnote}{-1}%
  \endgroup
}

\begin{document}

\title{A Review of Robot Learning for Manipulation: \\Challenges, Representations, and Algorithms}

\author{\name Oliver~Kroemer* \email okroemer@andrew.cmu.edu \\
       \addr School of Computer Science\\
       Carnegie Mellon University\\
       Pittsburgh, PA 15213, USA
       \AND
       \name Scott~Niekum* \email sniekum@cs.utexas.edu \\
       \addr Department of Computer Science\\
       The University of Texas at Austin\\
       Austin, TX 78712, USA
       \AND
       \name George Konidaris \email gdk@cs.brown.edu \\
       \addr Department of Computer Science\\
       Brown University\\
       Providence, RI 02912, USA\\
       }

\editor{Daniel Lee}

\maketitle

\begin{abstract}%
A key challenge in intelligent robotics is creating robots that are capable of directly interacting with the world around them to achieve their goals. The last decade has seen substantial growth in research on the problem of robot manipulation, which aims to exploit the increasing availability of affordable robot arms and grippers to create robots capable of directly interacting with the world to achieve their goals. Learning will be central to such autonomous systems, as the real world contains too much variation for a robot to expect to have an accurate model of its environment,  the objects in it, or the skills required to manipulate them, in advance. We aim to survey a representative subset of that research which uses machine learning for manipulation. We describe a formalization of the robot manipulation learning problem that synthesizes existing research into a single coherent framework and highlight the many remaining research opportunities and challenges. 
\end{abstract}


\section{Introduction}
\blfootnote{* Oliver Kroemer and Scott Niekum provided equal contributions.}\noindent Robot manipulation is central to achieving the promise of robotics---the very definition
of a robot requires that it has actuators, which it can use to effect change on the world. The potential for autonomous manipulation applications is immense: robots capable of manipulating their environment
could be deployed in hospitals, elder- and child-care, factories, outer space, restaurants, service industries, and the home.
This wide variety of deployment scenarios, and the pervasive and unsystematic environmental variations in even
quite specialized scenarios like food preparation, suggest that an effective  manipulation robot must be capable of
dealing with environments that neither it nor its designers have foreseen or encountered before. 

\begin{figure*}
\centering
  \includegraphics[width=0.775\columnwidth]{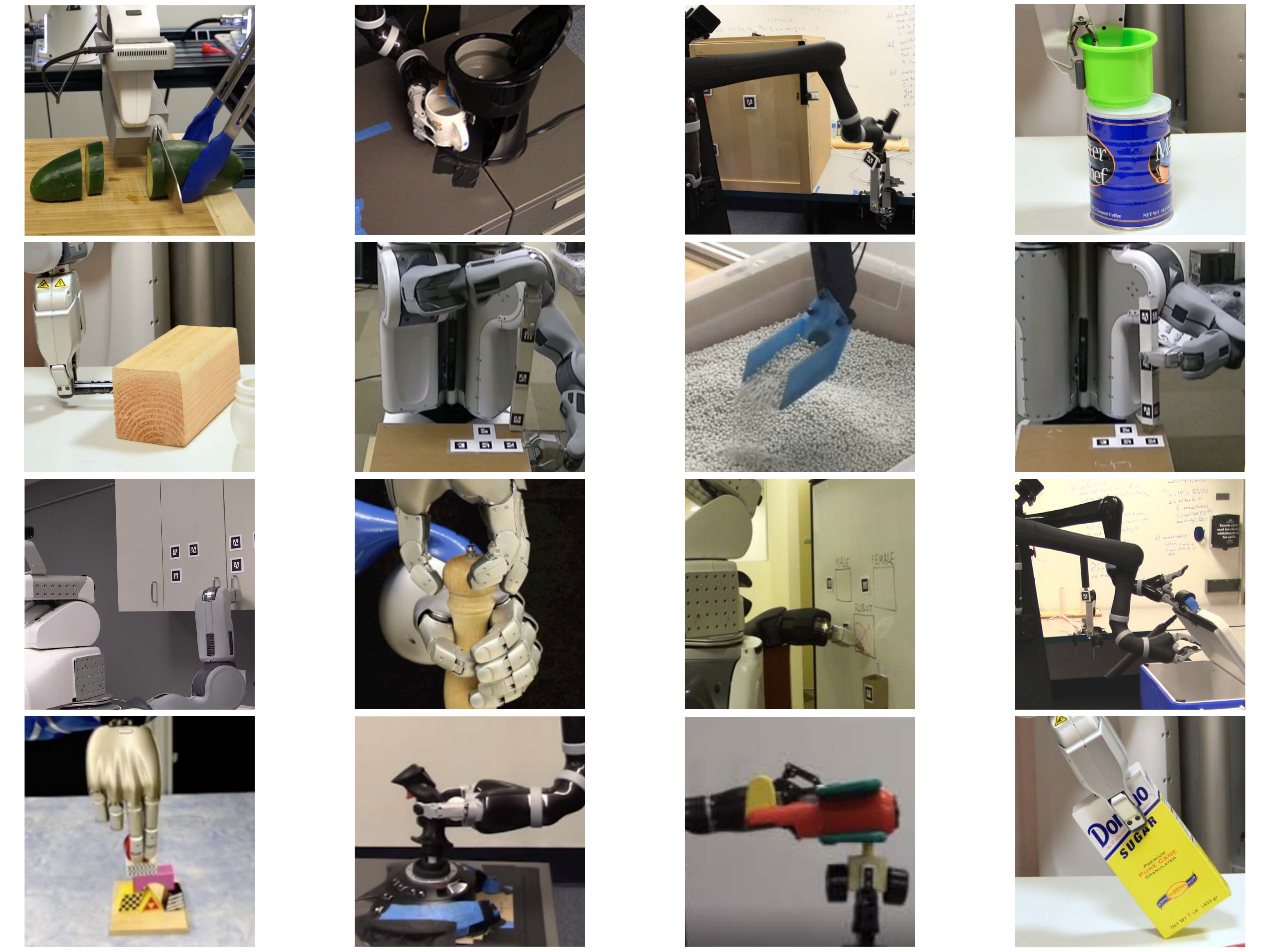}
  \caption{Example manipulation skills including inserting, stacking, opening, pushing, cutting, screwing, pouring, and writing.}
  \label{IntroFig}
\end{figure*}

Researchers have therefore
focused on the question of \textit{how a robot should learn to manipulate the world around it.} That research has ranged from learning 
individual manipulation skills from human demonstration, 
to learning abstract descriptions of
a manipulation task suitable for high-level planning, to discovering 
an object's functionality by interacting with it, and many objectives
in between. Some examples from our own work are shown in Figure~\ref{IntroFig}.

Our goal in this paper is twofold. First, we describe a formalization of the robot manipulation learning problem that synthesizes existing research into a single coherent framework. Second, we aim to describe a representative subset of the research that has so far been carried out on robot learning for manipulation. In so doing, we highlight the diversity of the manipulation learning problems that these methods have been applied to as well as identify the many research opportunities and challenges that remain.

\begin{figure}
\centering
\includegraphics[width=.75\textwidth]{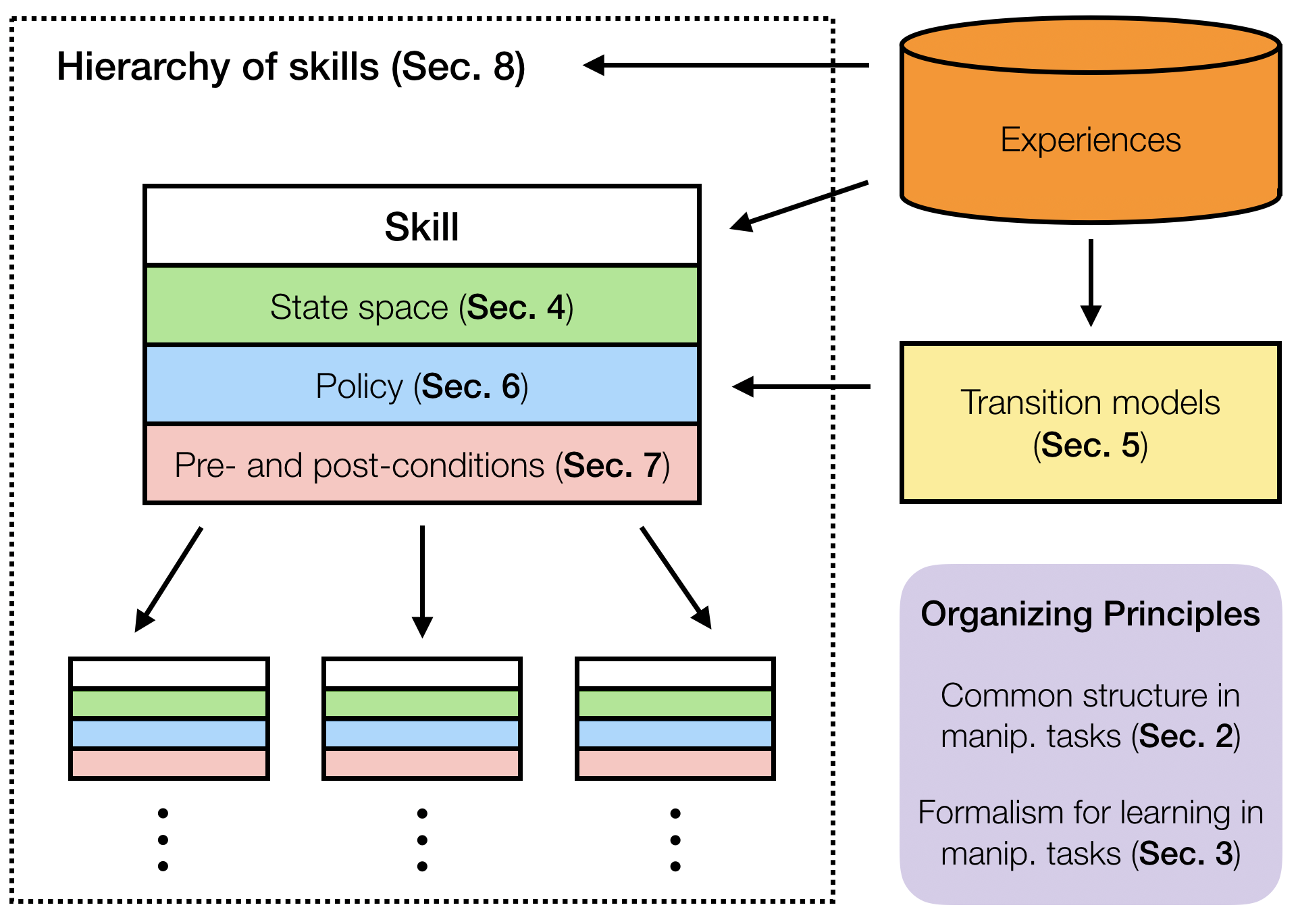}
\caption{Overview of the structure of the review.}
\label{fig:overview}
\end{figure}

Our review is structured as follows. First, we survey the key concepts that run through manipulation learning, which provide its essential structure (Sec. \ref{Section:ConceptsCommon}). Section \ref{sec:formalizing}  provides a broad formalization of manipulation learning tasks that encompasses most manipulation problems but which contains the structure essential to the problem.

The remainder of the review covers several broad technical challenges. Section \ref{sec:objects} considers the question of 
learning to define the state space, where the robot must discover the relevant state features and degrees of freedom attached to each object in its environment. Section \ref{sec:transitions} describes approaches to learning an environmental transition model that describes how a robot's actions affect the task state. 
Section \ref{sec:skills} focuses on how a robot can learn a motor control policy that directly achieves some goal, typically via  reinforcement learning \cite{sutton1998reinforcement}, either as a complete solution to a task or as a component of that solution. Section \ref{sec:prepost} describes approaches that characterize a motor skill, by learning a description of the circumstances under which it can be successfully executed, and a model of the resulting state change.
Finally, Section \ref{sec:hierarchy} surveys approaches to learning  procedural and state abstractions that enable effective high-level learning, planning, and transfer.

\section{Common Concepts in Learning for Manipulation}
\label{Section:ConceptsCommon}

Manipulation tasks have significant internal structure and exploiting
this structure may prove key to efficient and effective manipulation learning. Hence, before formalizing the manipulation learning problem, we will first discuss this internal structure.

\subsection{Manipulations as Physical Systems}
Every manipulation involves a physical robot interacting with its environment. As a result, all manipulations are subject to the laws of physics and the structure that they impose. This fairly obvious statement has wide-reaching implications for manipulation learning algorithms. Basic physical concepts (e.g., distinct objects cannot occupy the same space, and gravity applies a mass-dependent force to objects) provide strong prior knowledge for manipulation tasks. Concepts from physics, such as irreversible processes and object masses, are so fundamental that we generally take them for granted. However, these concepts provide invaluable prior knowledge and structure that can be exploited by learning algorithms and thus make learning manipulation skills tractable. Most of the concepts discussed in the remainder of this section are to some extent the result of manipulations being physical processes.

\subsection{Underactuation, Nonholonomic Constraints, and Modes in Manipulations}
Manipulation tasks are almost always characterized as underactuated systems. Even if the robot is fully actuated, inanimate objects in the environment will contribute a set of independent state variables to the state space, but not increase the robot's action space. The resulting discrepancy between the number of actuators and the number of DoFs means that the system is underactuated. To change the state of the objects, the robot must first move into a state from which it can alter the object's state and then apply the desired manipulation, e.g., make contact with an object and then push the object. These requirements can be represented as a set of nonholonomic constraints that define how the robot can move through the full state space based on different interactions with the environment. 

Manipulation tasks can be modelled as hybrid systems, wherein the system dynamics are continuous within each of a number of discrete dynamical modes. The dynamics are thus piecewise continuous. In manipulation tasks, the mode switches often correspond to making or breaking of contacts, with different contacts applying corresponding constraints and allowing the robot to interact with various objects. The conditions for transition between modes often correspond to subgoals or obstacles depending on which state variables the robot should change or keep constant. Unfortunately, modes also make manipulation tasks inherently discontinuous. Hence, small changes in the state can have a significant effect on the outcome of a manipulation. It is therefore important that robots monitor their skill executions for unexpected and undesired mode transitions.

\subsection{Interactive Perception and Verification}
Robots can perceive certain latent object properties by observing the outcomes of different manipulation actions. This process is known as \emph{interactive perception}. 
Many properties of objects, such as material or kinematic constraints, can only be determined reliably through interactive perception. If the goal of the task is to alter a latent property, then the robot will need to use interactive perception to verify that the manipulation was successful, e.g., pulling on a door to ensure it was locked. Even if a property's value can be approximated using passive perception, interactive perception often provides a more accurate estimate. In some cases, the estimate from the interactive perception can be used as the ground truth value for learning the passive perception. Interactive perception thus provides the basis for \emph{self-supervised learning}.
As the perception depends on the action, interactive perception is often combined with \emph{active learning} to actively select actions that maximize learning performance.

\subsection{Hierarchical Task Decompositions and Skill Reusability}
Manipulation tasks exhibit a highly hierarchical structure. For example, the task of cleaning a house can be divided into subtasks, such as cleaning the dishes, vaccuuming the floors, and disposing the trash. These subtasks can then be further divided into smaller subtasks, such as grasping a plate or trashbag. Even basic skills, such as grasping, can be further divided into multiple goal-oriented action phases. This hierarchy divides the primary task into smaller, more tractable problems. The robot can learn skill policies for performing the lowest level tasks and then use these skills as an action basis for performing the next level of tasks. The robot can thus incrementally learn a hierarchical policy of skills, with the resulting policy hierarchy reflecting the task hierarchy. The complexity of the learning challenges at each level of the hierarchy is reduced, enabling faster skill learning.

The hierarchy is also important because it results in a modular structure.  Subtasks and skills can often be interchanged to perform tasks in different ways depending on the scenario. Modularity also allows for some components to be predefined and others learned, e.g., an agent may be provided with a basic grasping reflex. More importantly, similar tasks will often appear multiple times within the hierarchy. For example, when cutting vegetables, each slice of the vegetable is a separate and slightly different task. These tasks are however similar enough that the robot should generalize across them rather than treat them as distinct with unique skills. We refer to such sets of similar tasks as \emph{task families}. By exploiting the similarity of these tasks, the robot can efficiently learn skills across entire task families and can thus be reused multiple times. The ability to incorporate this modularity into the robot's controllers and models is conditioned on having a suitable task decomposition. Discovering this structure autonomously by the robot is thus a major topic for manipulation research.

\subsection{Object-Centric Generalization}
One common structural assumption for manipulation tasks is that the world is made of up objects and that the robot's goals will typically be to modify some aspect or attribute of a particular set of objects in the environment.
Consequently, \textit{generalization via objects}---both across different objects, and between similar (or identical) objects in different task instances---is a major aspect of learning to manipulate.  Object-centric representations of manipulation skills and task models are often sufficient to generalize across task instances, but generalizing across different objects will require both motor skills and object models that adapt to variations in object shape, properties, and appearance. In some cases this can be done implicitly---e.g., with a compliant gripper that automatically adapts its shape to that of an object during grasping.
One powerful approach to generalizing across objects is to 
find an abstract representation under which we can consider
a family of objects to be equivalent or identical, even though
they vary substantially at the pixel or feature level, and adapt accordingly.

\subsection{Discovering Novel Concepts and Structures}
Robots working in unstructured environments will often come across new types of objects. Performing new tasks with these objects may require adapting existing skills or learning entirely new skills. Learning in open-world environments is thus not just a matter of the robot filling in gaps in its knowledge base. Instead, the scope of the knowledge bases will continue to expand, sometimes in unforeseen ways.
 The ability to handle novel concepts is an important aspect of robot autonomy, as it allows robots to handle unforeseen situations. To operate efficiently, robots will need to be capable of generalizing and transferring knowledge from prior experiences to structure the learning processes for these new concepts. This transfer learning may require more abstract reasoning depending on the similarity of the new concept to previous ones. Fortunately, as explained in this section, manipulation tasks exhibit a significant amount of structure that an autonomous robot can leverage when learning manipulation tasks.

\section{Formalizing Manipulation Learning Tasks}
\label{sec:formalizing}
\label{sec:MDP}
Robot learning problems can typically be formulated as individual Markov Decision Processes (or MDPs),\footnote{We should note that the Partially-Observable Markov Decision Process (POMDP) \cite{kaelbling1998planning} is a more accurate characterization of most robot learning problems. While these have been employed in some cases \citep[for example]{hsiao2007grasping,Vien15a,Vien15b}, the difficulty of solving POMDPs, especially in the learning setting, means their usage here is so far uncommon.}  described as a tuple:
$$(S, A, R, T, \gamma),$$
where $S \subseteq \mathbb{R}^n$ is a set of states; $A \subseteq \mathbb{R}^m$ is a set of actions; $R(s, a, s')$ is a \textit{reward function}, expressing the immediate reward for executing action $a$ in state $s$ and transitioning to state $s'$; $T(s' | s, a)$ is a \textit{transition function}, giving the probability distribution over states $s'$ reached after executing action $a$ in state $s$; and $\gamma \in [0, 1]$ is a discount factor expressing the agent's preference for immediate over future rewards. 
Here the goal of learning is to find a control policy, $\pi$, that maps states to actions so as to  maximize the \textit{return}, or discounted sum of future rewards $\sum_{i=0}^\infty \gamma^i r_i$, for that specific problem. 

This common formulation captures a wide range of tasks, enabling researchers
to develop broadly applicable general-purpose learning algorithms.
However, as we have discussed in the previous section, the robot manipulation problem has more structure; a major task for robot manipulation researchers
is to identify and exploit that structure
to obtain faster learning and better generalization. Moreover, generalization is so central to manipulation learning that a multi-task formulation is needed.
We therefore model a manipulation learning task as a structured
collection of MDPs, 
which we call a \textit{task family}. Rather than
being asked to construct a policy to solve a single task (MDP), we  instead aim to learn a policy that generalizes across the entire task family. 

A task family is a distribution, $P(M)$, over MDPs, each of which is a \textit{task}. For example, a door-opening task family may model each distinct door as a separate task. Similarly, a vegetable-cutting task family may model each individual slice as a task.
The action space is determined by the
robot and remains the same across tasks, but 
each task may have its own state space and transition and reward functions.  The reward function is formulated as a robot-dependent background
cost function $C$ ---shared across the entire family---plus a
reward $G_i$ that is specific to that $i^{th}$ task:
$$R_i = G_i - C.$$

The state space of the $i^{th}$ task is written as:
$$S_i = S_r \times S_{e_i},$$
where $S_r$ is the state of the robot and $S_{e_i}$ is the state of the $i^{th}$ environment. $S_{e_i}$ can vary from raw pixels and sensor values, to a highly pre-processed collection of relevant task variables. Many task environments will consist of a collection of objects and the variables describing the aspects of those objects that are relevant to the task. It is therefore common to model the environment as a collection of object states, resulting in a
more structured  state space where $S_{e_i}$ is partly factorized into a collection of relevant object states:
$$S_{e_i} = S_{w_i} \times \Omega^i_1 \times ... \times \Omega^i_{k_i},$$
 where $S_{w_i}$ is the state of the general environment, $\Omega^i_j$ is the state of the $j$th relevant object in task $i$, and task $i$ contains $k_i$ objects. The number of relevant objects may vary across, and occasionally within, individual tasks.
 Modeling tasks in this factorized way facilitates object-centric generalization, because policies and models defined over individual objects (or small collections of objects) can be reused in new environments containing similar objects. 
  This factorization can be clearly seen in symbolic state representations, wherein the modularity of proposition-based (e.g., \texttt{CupOnTable=True}) or predicate-based (e.g., \texttt{On(Cup,Table)=True}) representations allows the robot to consider only subsets of symbols for any given task.
  For manipulation tasks, we often employ predicate-based representations for their explicit generalization over objects. 
 
 Manipulation tasks also present modularity in their transition functions, i.e., the robot will only be able to affect a subset of objects and state variables from any given state. To capture the underactuated nature of manipulation tasks, we can model the tasks as hybrid systems with piecewise continuous dynamics. Each of the continuous dynamical subsystems is referred to as a \emph{mode}, and the state will often contain discrete variables to capture the current mode. Mode switches occur when the robot enters certain sets of states known as guard regions, e.g., when the robot makes or breaks contact with an object. The robot can thus limit the state variables that it may alter by restricting itself to certain modes.

 In some cases there is also structure in the action space, exploited through the use of higher-level actions, often called \textit{skills}. 
 Such skills are typically modeled using the options framework \cite{Sutton99}, a hierarchical learning framework that models each motor skill as an \textit{option}, $o = \left(I_o, \beta_o, \pi_o\right)$,
where:
\begin{itemize} 
\item $I_o: S \rightarrow \{0, 1\}$ is the \textit{initiation set}, an indicator function
describing the states from which the option may be executed. 
\item $\beta_o: S \rightarrow \left[0, 1\right]$ is the \textit{termination condition}, describing the probability that an option ceases execution upon reaching state $s$. 
\item $\pi_o$ is the option policy, mapping states in the option's initiation set to low-level motor actions, and corresponding to the motor skill controller. 
\end{itemize}
The robot may sometimes be pre-equipped with a set of motor skills that it can reuse across the family of tasks; in other settings, the robot discovers reusable skills as part of its learning process.

One of the key challenges of learning a policy, or domain knowledge,  across a task family is the need to transfer information from individual tasks to the whole family. A robot can learn the policy to solve a single task as a function of the task state. However, transferring these functions across the family is not trivial as the tasks may not share the same state space.  Transfer may be aided by adding extra information to the task---for example, information about the color and shape of various objects in the task---but since that information does not change over the course of a task execution, it does  not properly belong in the state space.
We model this extra information as a \textit{context vector} $\tau$ that accompanies each task MDP, and which the robot can use to inform its behavior. Like the state space, the context can be monolithic for each task or factored into object contexts. To generalize across a task family, the robot will often have to learn policies and models as functions of the information in the context vector. For example, a cutting skill needs to be adapt to material properties of different vegetable, or a door-opening skill needs to adapt to the masses and sizes of different doors.

In summary, our model of a family of manipulation tasks therefore consists of a \textit{task family} specified by a distribution of  manipulation MDPs, $P(M)$.
Each manipulation MDP $M_i$ is defined by a tuple
 $M_i = (S_i, A, R_i, T_i, \gamma, \tau_i)$, where: 
   \begin{itemize}
       \item   $S_i = S_r \times S_{e_i}$, is the state space, where $S_r$ describes the robot state, and $S_{e_i}$ the environment state. Often the environment state is primarily factored into a collection of object states: $S_{e_i} = S_{w_i} \times \Omega^i_1 \times ... \times \Omega^i_{k_i}$, for a task with $k_i$ objects; 
  \item $A$ is the action space, common across tasks, which may include both low-level primitive actions and a collection of options $O$;
  \item $R_i = G_i - C$ is the reward function, comprising a background cost function $C$ (common to
  the family) and
  a task-specific goal function $G_i$;
  \item $T_i$ is the transition function, which may contain exploitable structure across the sequence of tasks, especially object-centric structure;
  \item $\gamma$ is a discount factor, and
  \item $\tau_i$ is a vector of real numbers describing
 task-specific context information, possibly factored into object context: $\tau_i = \tau^i_1 \times ... \times \tau^i_k$, for $k$ objects.
   \end{itemize}

\begin{figure}
    \centering
    \includegraphics[width=\columnwidth]{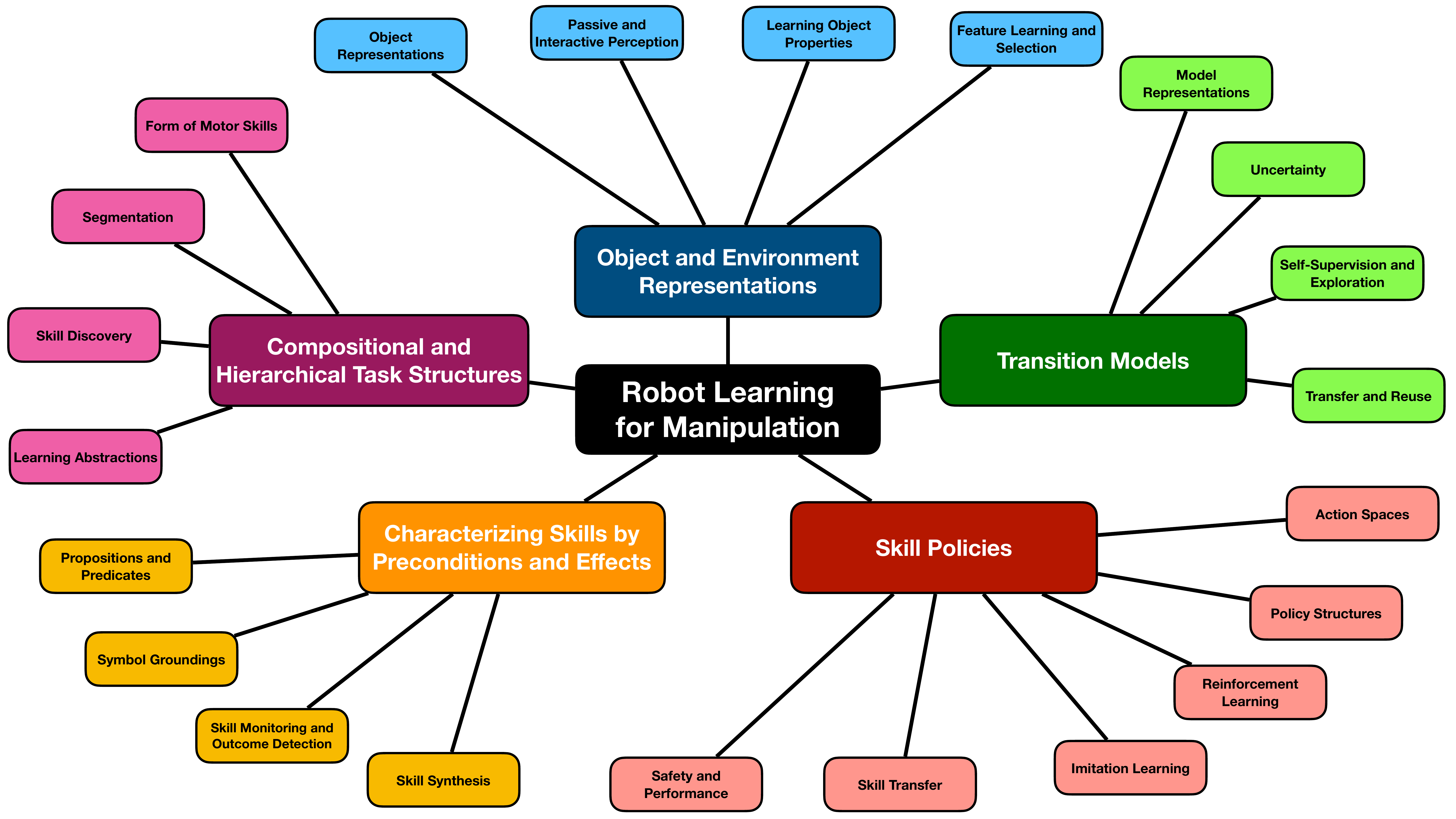}
    \caption{Overview of the different topics covered in this review sections.}
    \label{fig:tableofcontents}
\end{figure}

\textbf{Overview of Learning Problems for Manipulation: }
The learning problems posed by manipulation tasks can typically be placed into one of five broad categories, which we will discuss in the subsequent sections. An overview of the topics covered in this review are shown in Fig.~\ref{fig:tableofcontents}. 

When \textbf{learning to define the state space} (Sec. \ref{sec:objects}), the robot must discover the
    state features and degrees of freedom attached to each object in its environment. This information is assumed to be given in the traditional reinforcement learning
    and planning settings. That is not the case in robotics,
    and in particular in learning for manipulation, which involves interacting with
    objects that the robot's designers do not have a priori access to. Learned  representations  of object states 
    can be transferred across the task family as 
    components of each task's 
    state space.
    
When \textbf{learning a transition model of the environment} (Sec. \ref{sec:transitions}), the robot must learn a model of how its actions affect the task
state, and the resulting background cost, for use in planning. This is closely connected to learning to
define the state space.  If the learned transition models and reward functions are object-centric,
then they can be ported across the task family, resulting in a natural
means of object-centric generalization across tasks. 

When \textbf{learning motor skills} (Sec. \ref{sec:skills}), the robot attempts to learn a motor control policy that
directly achieves some goal, typically via  reinforcement learning \cite{sutton1998reinforcement}. 
Here, the goal ranges from learning task-specific solution policies, to policies 
that can produce a solution to any task in the task family given the context vector, to 
useful motor skills that constitute a component of the solution policy but are not themselves a complete solution.

Given a learned motor skill, we may also \textbf{learn to characterize that motor skill} (Sec. \ref{sec:prepost}), where the robot learns a description of the circumstances under which it can be successfully executed (often called preconditions, and corresponding to an option's initiation set $I_o$), and a model of the resulting state change (often called an effect).

Finally, \textbf{learning compositional and hierarchical structure} (Sec. \ref{sec:hierarchy}) aims to learn hierarchical knowledge that enables the
robot to become more effective at solving new tasks in the family. Here, the goal is to learn
component motor skills---completely specified options---and models of their operation to construct more abstract representations of the learning task.

\section{Learning Object and Environment Representations\label{sec:objects}}
Modelling manipulation tasks and generalizing manipulation skills requires representations of the robot's environment and the objects that it is manipulating. These representations  serve as the basis for learning transition models, skill policies, and skill pre- and post-conditions, as discussed in later sections. 

This section explains how the object-based state and context spaces of manipulation tasks can be defined and learned. We will also explain how the robot can  discover objects and estimate their properties using passive and interactive perception. 
Since many of the extracted object properties and features may be irrelevant to learning certain components of manipulation tasks, we conclude the section by discussing how to select and learn relevant features.

\subsection{Object Representations}

As discussed in Section \ref{Section:ConceptsCommon}, a robot's 
physical environment has  considerable  structure that it can exploit. In particular, the world can be divided into objects, each of which can be described by a collection of features or properties.  Examples include movable objects such as mugs, tables, and doors, and stationary objects such as counters and walls.  The robot can  create a modular representation by segmenting the environment into objects and then estimating the values of their properties. 
This representation supports the reuse of skills by allowing the robot to efficiently generalize between similar objects across different tasks.

Object representations capture how objects vary both within tasks and across tasks of the same family. Within-task variations are captured by the \emph{state} space---those features that a manipulation action can change; across-task variations are captured as part of the \emph{context} space---attributes that are fixed in any specific task but aid generalization across pairs of tasks. For example, when stacking assorted blocks, the shapes and sizes of the blocks are fixed for a given task and are thus part of the context. Different stacking tasks may however use different sets of blocks, and thus the context changes across the tasks.  Generalizing manipulation skills usually implies adapting, implicitly or explicitly, to variations in both context and state. For example, a versatile pick-and-place skill should generalize over the shapes of different objects (fixed in any specific task) as well as their positions in the environment (modifiable by actions during a task). 

\subsubsection{Types of Object Variations}

Several types of within- and across-task object variations are common in the literature. \textbf{Object pose} \cite{pastor2009learning,levine2013guided,Deisenroth2015} variations are the most common, and must be included in the state space when they can be manipulated (e.g., by a pick-and-place skill). However, in some cases, these can be fixed within a task but vary across the task family and thus belong in the context \cite{da2014learning,Kober:2011:Darts,JiangLearnToPlace,levine2013guided} (e.g., the height of a surface in a wiping task \cite{Do2014}. \textbf{Object shape} may vary within a task, via articulated \cite{Niekum2015b,Sturm,Katz2010,Martin-Martin2016,Sturm2011},
deformable \cite{Schenck2017Granular,Li_ICRA2016,SchulmanHLA13,Schenck2018,Li2018,seita2018robot,li2018learning}, or divisible \cite{Lenz2015,Worgotter2013,Yamaguchi2016} objects. Rigid object shapes may also vary across tasks \cite{BurchfielK18,Brandl2014Humanoids}, offering both a challenge to and opportunity for generalization \cite{Schenck2018}. Similarly, objects may vary in \textbf{material properties}, which can have significant effects on manipulation but typically only vary across-tasks (e.g., cutting objects of different materials  \cite{Lenz2015}), though there is some work on altering the material properties of manipulated objects \cite{chitta10tro,Schenck2012b,Isola2015}.

Finally, objects may vary in their \textbf{interactions or relative properties}, including both robot-object \cite{Bekiroglu2013,Kopicki} and object-object \cite{Stoytchev2005,Sturm,KroemerAuRo17Updated2,jund2018optimization} interactions. 
Objects may interact with each other resulting on constraints between them \cite{Jain2013}, and manipulation may result in mode switches that add or remove constraints \cite{Niekum2015b,17-baum-Humanoids,Kroemer2014,Baisero2015}. Variance
across, rather than within, tasks is also possible---e.g., the properties of the joint connecting a cabinet and its door will remain constant for a given cabinet, but different cabinets can have different constraints \cite{Sturm,Hausman2015,Niekum2015b,Dang2010}.

\subsubsection{Object Representation Hierarchies}

Object models can be represented hierarchically, with layers corresponding to point-, part-, and object-level representations, each decreasing in detail and increasing in abstraction, and affording different types of generalization.  
This representation hierarchy mirrors the geometric structure of objects and their parts. Geometric properties and features capture \emph{where} the points, parts, and objects are, and non-geometric properties tend to capture the corresponding \emph{what} information defining these elements. Representations at each level of the hierarchy can capture general inherent properties as well as semantic or task-oriented properties \cite{DangSemGrasp,MyersAffordanceSegmentation,JangSemGrasp}. In addition to representing individual objects, the robot can also represent interactions between objects at the different layers of the hierarchy.
 
\textbf{Point-level Representations:} Point-level representations are 
 the lowest level of the hierarchy, and include point cloud, pixel, and voxel representations for capturing  the partial or complete shapes of objects in detail  \cite{Schenck2017Granular,Bohg2010,Klingensmith-2014-7883,Florence2018,Choi18}.  
Point-level representations provide the robot with the most flexible representations for capturing important details of objects and manipulation tasks. 
 
 Each element of these representations may be associated with additional features, such as the color or material properties corresponding to this point. Segmentation methods can be used to assign labels to individual points according to which part or object they belong \cite{Schenck2018,MyersAffordanceSegmentation}. Interactions may be modeled at this level of the hierarchy as contact points \cite{KaulOWA16,DangGraspStability,RosmanSpatial,Kopicki,Vina,Su,VeiHofPetHer2015,Piacenza2017}.
 
 Generalization across tasks can be accomplished by establishing correspondences 
between objects and environments at the point level. The robot may identify individual key points of objects (e.g., the tip of a tool \cite{Edsinger}),  or use non-rigid registration or geometry warping  to determine correspondences of entire sets of points across task instances \cite{SchulmanHLA13,Hillenbrand2012,Stuckler2014,Rodriguez2018,amor2012}. These correspondences can then be used to directly map manipulation skills between tasks or to compute higher-level representations \cite{Stuckler2014}.

\textbf{Part-level Representations:} 
Representations at the part level correspond to sets of multiple contiguous points from the lower level of the hierarchy \cite{sung_dme_2017,detry2013a,Dang2010,Byravan2016}, typically  focusing on parts associated with certain types of manipulations. For example, a mug can be seen as having an opening for pouring, a bowl for containing, a handle for grasping, and a bottom for placing \cite{FAGG19981277}. Each part can then be described by a set of features describing aspects such as shape, pose, color, material, type, and surface properties. 
Robots can use part-level features to represent interactions or relations between the parts of different objects, or to indicate the types of interactions and constraints imposed between the parts due to the interaction (e.g., a peg must be smaller than the hole for an insertion task).  

Defining correspondences at the part level enables generalization across different types of objects \cite{tenorth13cad,sung_robobarista_2015,detry2013a}. Many objects  have similar parts that afford similar interactions, although the objects may be very different as a whole. For example, a coin and a screwdriver are distinct types of objects, but they both have short thin metallic edge that can be used to turn a screw. Similarly, many objects have graspable handles; identifying the handle correspondences thus allows for transfering  grasping skills \cite{detry2013a,Kroemer2012}. 
Part-based representations therefore allow the robot to generalize between different classes of objects without having to reason about individual points of the object \cite{sung_robobarista_2015}. 

\textbf{Object-level Representations:}  
Object-level representations are important because robots generally select objects, not individual features, to manipulate \cite{Janner2019,Deisenroth2015,Gao16-ICRA-Deep,Jang2018}. Thus, robots must 
generalize between the sets of features attached to each object. Useful object-level representations group together 
object-specific properties such as an object's pose, mass, overall shape, and material properties (for uniform objects). Semantic object labels can be used to distinguish different types of objects and how they should be manipulated \cite{JangSemGrasp}. 
Similar to parts, object-level interaction features often define the relative poses, forces, and constraints between objects in manipulation tasks, e.g., relative poses for stacking blocks \cite{fang16simcontrol,Jain2013,Ugur2015,Sundaralingam2019}. The robot may also define different types of interactions at this level (e.g., object A is \emph{on} or \emph{inside} object B) or relations between objects (e.g., relative sizes and weights) to capture sets of interactions in a more abstract form \cite{Schenck2012,Schenck2018,Kulick2013,Schenck2012b,Fichtl}. 
Generalization across objects typically requires establishing correspondences between distinct
objects \cite{DevinADL18} that support similar manipulations.

Robots may also need to represent \textbf{groups of objects}---rather than modeling individual objects within the group, it is often more efficient and robust to use features that represent groups of objects as a whole entity. For example, clutter, or the background scene with which the robot should avoid colliding, is often represented in a similar manner to a single deformable or piece-wise rigid object. Identifying and reasoning about individual objects may not be necessary, and may even add additional complexity to  learning, causing learning to generalize poorly or be less robust.   Manipulating a specific object among a group  may require the robot  to recognize it, or even actively singulate the object \cite{Gupta0S15,Hermans2012}.

\subsection{Passive and Interactive Perception}
As embodied agents capable of manipulating their surroundings, robots can use actions to enhance their perception of the environment. Robot perception is therefore broadly divided into passive and interactive perception, with the key difference being whether or not the robot physically interacts with the environment.

The term \textbf{passive perception} refers to the process of perceiving the environment without exploiting physical interactions with it, i.e., non-interactive perception \cite{Isola2015}---for example, recognizing and localizing objects in a scene based on a camera image \cite{BurchfielK18,Tremblay2018DeepOP,wang2019,Yamaguchi16}.  Passive perception allows the robot to quickly acquire a large amount of information from the environment with little effort. 
Passive perception does not require the environment or the sensor to be stationary; observing a human performing a manipulation task is still  passive, as the robot does not itself perform the interaction. Similarly, moving a camera to a better vantage point is still a non-interactive form of perception, as the robot does not apply force to, or otherwise alter, the state of the environment\cite{saran2017viewpoint,Kahn}.

In \textbf{interactive perception} \cite{BohgInteractivePercep}, the robot physically interacts with its surroundings to obtain a better estimate of the environment. For example, a robot may push an object to better estimate its constraints or lift an object to estimate its weight \cite{Barragan2014,Hausman2015,Katz2011}. Robots can use a wide range of sensor modalities to observe the effects of its interactions, including haptic, tactile, vision, and audio \cite{lenz2015deepmpc,chitta10tro,Hogman2016,Griffith,thomason2016learning}. 

The need to perform actions means that interactive perception requires more time and energy than passive perception. The benefit of interactive perception is its ability to disambiguate between scenarios and observe otherwise latent properties \cite{TsikosB91,chitta10tro,Gao16-ICRA-Deep}, enabling the robot to reduce uncertainty. For example, the robot may not know whether two objects are rigidly connected or simply in contact; interactive perception allows it to test each hypothesis. 

Different actions will result in different effects and hence robots can learn about their environment faster by selecting   more informative actions \cite{Barragan2014,17-baum-Humanoids,Dragiev2013,Otte2014,Kulick2015,saran2017viewpoint,Kenney2009}. For example, shaking a container will usually provide more information regarding its contents than a pushing action would \cite{Schenck2012b,Sinapov2014,Schenck2014}.
Active learning approaches usually estimate the uncertainty of one or more variables in the environment and then select actions based on the resulting entropy, information gain, or mutual information \cite{Otte2014,Hogman2016,Hausman2015,VanHoof2014,Kulick2015}. 

The ability to test hypothesis means that interactive perception can also be used as a supervisory signal for learning to estimate properties using passive perception \cite{Griffith,PintoCuriousRobot,VanHoof2014,Nguyen2014,Wu2016,Kraft2008,pathak2018}. As an example of interactive learning, a robot may learn to predict the mass of objects from their appearance by first using interactive perception to determine the masses of a set of training objects. This form of self-supervised learning allows the robot to gather information autonomously and is thus crucial for enabling robots to operate in unfamiliar environments.

\subsection{Learning About Objects and Their Properties}
Having explained the different types of object variations and types of perception, we now discuss how the robot can learn about the objects around it from data.

\subsubsection{Discovering Objects}
A common first step in learning is distinguishing the individual objects in the scene, which is a segmentation problem, typically accomplished using passive perception \cite{Kraft2008,AlexeObjectness,Schwarz2018,Byravan2016,he2017maskrcnn}.
 However,  objects may often be close together in the scene, presenting the robot  with ambiguous information about object identity. Here the robot can maintain a probabilistic belief over whether or not different parts belong to the same object \cite{VanHoof2014,Kenney2009,hausman12interactive},
 and use interactive perception or viewpoint selection to disambiguate the scene \cite{Gualtieri2017,Hermans2012,VanHoof2014,hausman12interactive}.

 \subsubsection{Discovering Degrees of Freedom} 
Once individual objects have been identified, the robot may need to identify their kinematic degrees of freedom \cite{Niekum2015b,jain2019learning,hausman12interactive,Katz2010,17-baum-Humanoids,Sturm,Katz2011,abbatematteo2019learning}. These constraints and articulated connections are fundamental to establishing the objects' state space for manipulation tasks, as well as for robust pose tracking \cite{desingh2018factored,SchmidtNF15}. The different types of joints are usually represented using distinct articulation models, e.g., revolute or prismatic, with their own sets of parameters. For example,  a revolute joint model is specified by the position, direction, and limits of its rotation axis \cite{Sturm,Barragan2014,Niekum2015b,Sturm2011}. The robot must estimate these context parameters to accurately model the degrees of freedom. 
Passive perception can be used to estimate the kinematic chains of articulated objects, especially if the object is being moved by a person \cite{Martin-Martin2016,Niekum2015b,pillai14,Brookshire,jain2019learning}. Interactive perception is also well suited for estimating the articulation model parameters and the resulting motion constraints \cite{Sturm,Barragan2014,Hausman2015}.  Given the high-dimensional parameter space of the articulation models, active learning approaches are often used to select informative actions for quickly determining the model parameters \cite{Hausman2015,17-baum-Humanoids}, or transfer learning methods may be used \cite{abbatematteo2019learning}.

\subsubsection{Estimating Object Properties}
Once the robot has identified an object in the environment, the next step is to estimate
the object's properties. Since  some properties are only applicable
to some classes of objects \cite{diuk2008object,Bullard2018,Dragiev2013,Bhattacharjee2015,Schenck2012b,Wu2016,chitta10tro} 
the robot must first recognize the object class. For manipulation tasks, object classes are often derived from the actions and interactions that they afford the robot, e.g., container, graspable, and stackable, such that the interaction-relevant properties can be easily associated with the objects. 

Here the robot may use passive and interactive perception to estimate object property values \cite{Gao16-ICRA-Deep,Isola2015,Li2014,Varley2018}. In addition to object recognition, passive perception is often used to estimate the position, shape, and material properties \cite{Li2018,Bhattacharjee2015,Wu2016,Tremblay2018DeepOP,Cifuentes,SchmidtNF15,BurchfielK17,Jan2016,BurchfielK18,wuthrich2015,Yamaguchi16}. However, additional interactive perception can often be used to acquire more accurate estimates of these properties \cite{Chitta2010,BjorkmanTactileGlances,Dragiev2013,Javdani-2013-7698,Petrovskaya,Koval2017}.
Material and interaction properties are often difficult to estimate accurately using only passive perception. The estimates can be significantly improved by interacting with the objects and using tactile sensing \cite{Sinapov,SinapovVibrotactile,Sinapov2014,Sung2017,Gao16-ICRA-Deep,lenz2015deepmpc,HsiaoRSS10,Varley2018}. Exploratory actions can also be used to estimate the dynamic properties of objects, e.g., the center of mass, or the contents of containers \cite{Guler2014,Schenck2014,chitta10tro}.

\subsection{Feature Learning and Selection}
Even though an environment may contain a lot of objects and sensory stimuli, only a few object properties or sensory signals will usually be relevant for a given task. For example, when opening a bottle, the size of the cap is a relevant feature, but the color of the chair is irrelevant.
Using a suitable set of relevant features simplifies the skill and model learning problems. It also increases robustness and generalization to new situations. If the set of object properties is sufficiently rich, then the robot may only need to \emph{select} a suitable set of these as features for learning \cite{DevinADL18,KroemerMeta,montesano2008learning,Song2011,Stramandinoli2018}. In many cases, the robot will however need to \emph{learn} a set of features for the given task.

Unsupervised feature learning methods extract features from unlabeled training data. 
Dimensionality reduction methods may be used to capture correlations in data and discard noisy components of signals. 
For manipulation domains, dimensionality reduction approaches can be used to learn compact representations of complex objects or variations within classes of objects \cite{BurchfielK17,Bergstrom2012}.
Dimensionaly reduction methods can also be used to reduce the robot's action space of grasps to better fit common object shapes \cite{Ciocarlie2007}.
Clustering methods are used to cluster together similar data samples. For manipulations, clustering method may for example be used to cluster together distinct types of objects or continuous effects of manipulations \cite{Ugur2015}.

Supervised approaches learn features as part of the overall model or skill learning process. Decision trees and neural networks are often used to learn features in a supervised setting. Deep learning in particular has become ubiquitous for feature learning in robotics. Different network structures and layers, such as auto-encoders, spatial soft max layers, convolutions, and segmentation masks, can be used to incorporate architectural priors for learning useful features.
For example, segmentation masks can be used to capture the co-movement of nearby points in an image to capture points of objects or parts moving together \cite{Byravan2016,FinnL17}. 
Deep neural network models can be used to represent state spaces for manipulation tasks with high dimensional observation spaces. Representations of manipulation environments can be learned in a task-oriented manner to facilitate subsequent planning and control \cite{Kurutach,srinivas2018universal}. Neural networks are also highly effective at combining data from multiple sensor modalities or information sources. For manipulation tasks, robots often use this approach to merge information between passive modalities (e.g., vision) and more interactive modalities (e.g., touch and haptics), or to incorporate additional task information (e.g., instructions) \cite{Gao16-ICRA-Deep,lee2019icra,sung_dme_2017}.

\section{Learning Transition Models}
\label{sec:transitions}
\label{sec:modellearning}

The goal of manipulation tasks is to alter the states of objects in the robot's environment. Learning transition models that capture the changes in state as a result of the robot's actions is therefore an important component of manipulation learning. 

\subsection{Representing and Learning Transition Models}
Learning a transition model requires a suitable representation. The general form of a transition model for a family of MDPs is either  a deterministic function $T:S \times A\rightarrow S$ or a stochastic distribution $T:S \times A \times S \rightarrow \mathcal{R}$ over the next states given the current state and action. The transition models can also depend upon the context vector $\tau$ in order to explicitly generalize between contexts. A transition model is often employed iteratively to perform multi-step predictions.   

\textbf{Continuous Models:}
In manipulation tasks, robots are generally operating in continuous state and action spaces, e.g., the state is given by the continuous poses of objects and the actions are given by desired joint positions or torques. Robots therefore often learn low-level transition models for predicting the next state as a set of continuous values,
even when the set of actions is discrete \cite{Scholz2010}. Regression methods, such as neural networks, Gaussian processes, and (weighted) linear regressions, are commonly used for learning transition models \cite{Schenck2017Granular,levine2013guided,deisenroth2011pilco,Atkeson1997}. 

The simplest models, such as linear regression models, can be learned from small amounts of data \cite{Stilman2008}, but often underfit the inherent nonlinearities of most manipulation tasks. Hence, the generalization performance of these models is  limited. Instead, time-dependent linear models are often used to learn local models for specific regions of the state space \cite{Kumar2016}, e.g., around the trajectory of a skill being learned. These models may be sufficient while maintaining low data requirements, and can be combined with prior models that capture more global information about the task. Gaussian mixture models and neural networks have both been used to learn such informative priors \cite{fu2016one}.

Nonparametric models, including Gaussian processes and locally weighted regression, can flexibly learn detailed transition models \cite{Deisenroth2015,Kopicki2017}. The generalization performance of these types of models depends heavily on their hyperparameters and is often fairly limited. However, local models can be learned from relatively small amounts of data. For example, an accurate model for planar pushing of a block may be learned from less than a hundred samples \cite{Bauza}.

More complex hierarchical models, such as neural networks and decision trees, allow the robot to learn task-specific features that provide better generalization \cite{lenz2015deepmpc,Schenck2017Granular,fu2016one,Janner2019,gonzalez18a,FinnL17,Kurutach,srinivas2018universal,YeWDG18}.
For example, convolutional encoder-decoder networks can be used to learn models for predicting the transitions of individual points in the scene based on the selected actions \cite{Byravan2016,FinnL17}. However, unless the models are pre-trained, learning intermediate features significantly increases the amount of data required for training these models. Broadly speaking, the best representation for learning a continuous transition model often depends on the amount of training data available and the model expressiveness required by the task. 

Predictive state representations (PSRs) allow the robot to model the distributions over future observations based on the history of past observations \cite{Littman2001,Singh2004,Boots2013,Stork2015}. These models thus capture latent state information, as part of a POMDP formulation, without explicitly modeling or inferring the latent state. The robot can thus work directly with the observed variables.

More advanced parametric models provide additional structure, up to including entire physics engines \cite{Wu2015,Scholz,Wu2016,Li2015}. These engines provide a significant amount of prior structure and information, which improves generalization and data efficiency. However, even given the engine, it is difficult for a robot to map a real world scenario into a simulation environment. The simulation may also still fail to accurately capture certain interactions due to their lack of flexibility. Physics engines are therefore sometimes used to provide a prior or to initialize a more flexible representation by combining analytical models with flexible data-driven models \cite{ajayIROS2018}. 

\textbf{Discrete Models:}
Discrete transition models are used to learn transitions for tasks with discrete state and action spaces, typically capturing high-level tasks. For example, a robot may learn a discrete transition model for flipping a pancake, in which the transition probabilities indicate the probability of the pancake being flipped, dropped, or not transitioning. The model can then be used to predict the outcome of attempting multiple flips. Discrete models can also capture low-level transitions where the state space has been discretized. This results in the  continuous states corresponding to a single discrete state being considered as identical. It also removes the smoothness assumption between states, such that state transition information is not shared between discrete states. As a result,  a very fine-grained discretization will lead to poor generalization and a coarse discretization may not capture enough detailed transition information to perform the task. A suitable discretization should therefore cluster together states that will result in similar transitions given the same actions \cite{Bergstrom2012}. Adaptive or learned discretizations are a form of feature learning \cite{guan2018efficient}. 

The most basic representations for discrete models are tabular models or finite state machines. Discrete models treat each state as atomic, with a transition distribution independent of all other states. This approach however severely hampers generalization and transfer. 
More structured models use collections of symbols to represent discrete state spaces. Proposition-based models define a set of propositions \cite{konidaris2018skills}, e.g., \texttt{Obj1Grasped} and \texttt{Obj2Clear}, which are either true or false and thus define the state. A set of $N$ propositions thus defines a set of $2^N$ states. The factorization provided by the propositions results in a more compact representation of the state space, and also allows transition information to be shared between these states. In particular, symbolic models are often used to express compact (and sometimes stochastic) action models known as \textit{operators}. Each such operator has a a precondition  that must be true for the action to be executable, e.g., \texttt{WaterReservoirFull=True} and \texttt{CoffeeJugFull=False} for a coffee making action. The effects of the action are then given by lists of propositions that become true (e.g., \texttt{CoffeeJugFull}) and that become false (e.g., \texttt{WaterReservoirFull}), with all other propositions being unchanged by the transition. As the preconditions and effects only refer to a small number of propositions, the operator generalizes across the values of all other propositions that are not mentioned.

These models may be generalized to relational or first-order models \cite{Lang2012}, which support extremely compact action models and enable generalization across objects. 
This approach allows predicate-based models (e.g., \texttt{Full(CoffeeJug)} for a full jug of coffee and \texttt{Full(WaterReservoir)} for a full water reservoir)  to be used for families of tasks with different numbers of objects and states. This extended generalization can unfortunately also have some unintended effects if distinct situations are treated as the same. For example, a dishwasher loading action would treat a paper plate the same as a ceramic plate unless a suitable predicate, e.g. \texttt{IsCeramic(Plate)}, were included to avoid this type of over generalization.

\textbf{Hybrid Models:}
Hybrid models combine elements of discrete and continuous models, often resulting from the  hierarchical structure of manipulation skills and subtasks. The discrete components of the state are often used to capture high-level task information while the continuous components capture low-level state information. 

Robots will often learn continuous models for specific subtasks. For example, a robot may learn separate models for opening a door, moving its body through free space, and stacking newspapers. These models can then be used to predict the effects of actions within each skill. In turn, these models can then become a part of the robot's overall world model to predict the effects of sequences of skills. This approach allows the robot to exploit the modularity of the distinct subtasks for the discrete high-level transition model. 
Many works on learning continuous models for individual skills or subtasks can thus be seen as learning part of a larger hybrid model.

One common use of hybrid models is to represent modes in manipulation tasks \cite{HauserThesis,Simeon2004,Kroemer,Lee2017,Choudhury2017,Zhou2018}. In these cases, the dynamics are piece-wise continuous with jumps between the modes. These jumps are often defined by guard sets wherein the state transitions to a certain mode if the continuous state enters the guard set. In manipulation tasks, the jumps between modes  typically  result from contacts being made or broken between objects.
The mode switches thus capture the changing dynamics and constraints caused by the changing contacts. Planning approaches for manipulation domains often explicitly take this multi-modal structure into account for planning \cite{Toussaint2018,jain2018efficient}.
Learning a hybrid model entails learning both the dynamics within each mode and the conditions for switching between modes. The dynamics within a mode can be learned using standard continuous model approaches. The guard regions can be modeled as explicit sets of states or using classifiers. More flexible models, such as nonparametric and neural network models, may also be able to implicitly capture hybrid dynamics of manipulation tasks \cite{FinnL17}. Exploiting the modularity of they hybrid system is however more difficult when using an implicit model of the mode structure.

\subsection{Uncertainty in Transition Models}

State transitions are often represented using probabilistic models that  allow the robot to represent multiple potential outcomes. For example, a robot may choose to move to a trash can and drop an object inside because the resulting state distribution has a lower variance than attempting to throw the object into the trash. When working with probabilistic models, it is important to distinguish between two sources of uncertainty: aleatoric and epistemic uncertainty. 

\textbf{Aleatoric uncertainty:}
A stochastic process exhibits a randomness in its state transitions. For example, when throwing a block into the air, the side on which the block will land is random. Similarly, due to slight variations in the execution and surface properties, the exact location of an object being placed on a table may follow a Gaussian distribution. Attempting to open a door may only result in the door opening $98\%$ of the time and remaining closed the other $2\%$. 

Stochasticity resulting in aleatoric uncertainty is common in manipulation tasks, for both discrete and continuous state spaces \cite{Kopicki2017,Lang2012}.  Probabilistic models are thus used to capture the process noise and approximate the distributions over next states \cite{babaeizadeh2018,Kroemer}.

\textbf{Epistemic uncertainty:}
In addition to the inherent stochasticity of manipulation, the outcome of an action may be uncertain due to the robot's limited knowledge of the process. For example, consider the task of placing a closed container into a bowl of water. Whether the container will float or sink is deterministic given its mass and volume. However, the robot's prediction over the two outcomes may still be uncertain if the robot does not know the mass of the container or because it lacks sufficient training samples to accurately predict the mass threshold at which the container will sink. The probabilistic prediction is thus the result of epistemic uncertainty, or uncertainty over the model parameters, rather than aleatoric uncertainty from inherent stochasticity. Unlike aleatoric uncertainty, epistemic uncertainty can be reduced given additional information, e.g., using interactive perception to better estimate the mass of the container or acquiring additional training samples for estimating the threshold for the model. 

Bayesian approaches explicitly capture epistemic uncertainty in their predictions by representing the probability distributions over different potential models \cite{Deisenroth2015,Scholz}. They also incorporate explicit prior beliefs over these model distributions.  By contrast, point estimates, such as maximum likelihood or maximum a posteriori MAP estimates, assume a single model given the current data and use this model to predict the next state. Linear Bayesian models are often too simple to capture the complexity of manipulation tasks. Kernel-based Gaussian processes have therefore become one of the most popular Bayesian approaches for learning transition models \cite{Deisenroth2015,Hogman2016}. Gaussian processes, as well as linear Bayesian regression, both have explicit parameters modeling the stochasticity of the system being represented. The model uncertainty can also be used to guide exploration for learning an accurate model from fewer samples \cite{wangIROS2018}.

\subsection{Self-supervision and Exploration for Learning Transitions}

Transition models are usually learned in a self-supervised manner. Given the current state, the robot performs an action and observes the resulting effect on the state. The robot can thus acquire state, action, and next state tuples for training the model. To generalize more broadly, the robot will also need to estimate the context parameters for each task and then incorporate these into the model as well.

The robot can adopt different exploration strategies for acquiring samples. Random sampling is often used to learn general models and acquire a diverse set of samples. A grid approach may be used to ensure that the action samples are sufficiently spread out, but these approaches generally assume that the state can be reset between actions \cite{Kuan2016}. Active sampling approaches can be used to select action samples that are the most informative \cite{wangIROS2018}. If the model is being used to learn and improve a specific skill, the robot may collect samples within the vicinity of the skill using a small amount of random noise (see model-based skill learning in the next section). Another increasingly popular approach to exploring an environment is to use intrinsic motivation \cite{Chentanez2005,pathak2019}. In this case, the robot actively attempts to discover novel scenarios where its model currently performs poorly or that result in salient events.

\subsection{Transferring and Reusing Transition Models}
Transition models are not inherently linked to a specific task, and can therefore often be transferred and reused between different manipulation tasks and even task families.  For example, a model learned for pushing an object may be used as the basis for learning push grasps of objects. In order to directly reuse a model, the learned model and the new task must have the same state, action, and context spaces, or a mapping between the spaces may be necessary \cite{taylor2007trans}. Given compatible spaces, the ability to transfer or reuse models depends on the overlap between data distributions for training the old and new task models. This issue is known as covariate shift (input varies) and dataset shift (input and output varies). If the previous model was trained on data distinct from the needs of the new task, then the benefit of using the previous model will be limited and may even be detrimental. Assuming a sufficiently rich model, the robot can acquire additional data from the new task and use it to update the model for the new region. In this manner, the robot's learned model may become more applicable to other tasks in the future. Data can also be shared across models when learning skills in parallel.

\section{Learning Skill Policies} 
\label{sec:skills}
All of the robot learning problems discussed thus far are ultimately in service of helping the robot learn a policy that accomplishes some objective.	    
Thus, the final learning goal for the robot is to acquire a behavior, or \textit{skill controller}, that will perform a desired manipulation task. 
A common representation for skill controllers is a \textit{stochastic policy} that maps state-action pairs to probabilities (or probability densities in the continuous case).  
This section discusses the spectrum of different policy parameterizations that can be chosen, algorithms for learning skill policies from experience and demonstrations, methods for transferring knowledge across skills and tasks, and approaches that can provide safety and performance guarantees during learning.

\subsection{Types of Action Spaces}
The choice of action spaces is an important part of designing a manipulation policy. The robot ultimately needs to send a control signal to its actuators to perform actions in the physical world. These signals may, for example, define the desired pressure for a hydraulic or pneumatic actuator \cite{Bischoff,Gupta2016}, the heating for a shape memory alloy actuator \cite{AbuZaiter2016,Zimmer2019}, the tendon activation for a cable-driven robot \cite{Schlagenhauf2018,DexManip2020}, or the torque for an electric motor \cite{NGUYENTUONG2011,levine2015learning}. Robots that are difficult to model, such as soft robots and robots with complex dynamics, may often benefit from learning policies that directly output these control signals. However, the outputs of policies often do not work directly at the level of actuator signals.

In practice, an additional controller is often placed between the policy and the actuator \cite{Gullapalli}. Using an additional controller allows the robot to leverage a large body of prior work on control for robotics \cite{Spong}. Example controllers include simple linear PID controllers as well as more complex model-based admittance and impedance controllers. The policy's actions then define the desired values for the controller. The action space of the policy often defines the desired positions, velocities, or accelerations for movements or desired forces and torques for interactions.

Both desired position and force information can be defined in the joint space or a Cartesian task space for the end effector. For manipulation tasks, it is often easier to generalize interactions with objects across the robot's workspace using a Cartesian action space \cite{Mason1981,Ballard1984}. 
For example, with a Cartesian action space, applying an upward force on a grasped object would be the same action anywhere in the robot's workspace. The controller is then responsible for mapping these desired signals into the joint space for actuation.  For a joint-space action policy, the robot would need to learn different joint torques depending on the arm's current configuration.

Although the Cartesian actions could be defined in a global or robot-centric coordinate frame, additional generalization can often be achieved by using task frames associated to individual objects or salient features of the environment \cite{Ballard1984}. These task frames may be predefined, selected, or learned. Using a given object-relative task frame allows the policy learning to focus on how to perform the task rather than where to perform it. 

In addition to determining the desired values for inputs to the controller, the policy may also output additional values for adapting the controller. In particular, a policy may define different controller gains as part of its action space \cite{Buchli2011,Tianyu18}. In this manner, the policy can make a robot more compliant or stiffer at different points during the task execution. Otherwise the gains are usually predefined and fixed. 

The inclusion of a controller also allows the robot to use a policy that operates at a lower frequency than the controller. While the low-level controllers may operate at 100s or 1000s of Hertz, the policies can operate at lower frequencies. For policies operating at lower frequencies, an additional interpolation step may be used to guide the controller between the desired values.

\subsection{The Spectrum of Policy Structure}
In robotic manipulation, specific parameterizations are often used that restrict the representational power of the policy; if these restrictions respect the underlying structure of the task, generalization and data efficiency are often improved without significantly impacting asymptotic performance. Thus, the choice of policy representation is a critical design decision for any robot learning algorithm, as it dictates the class of behaviors that can be expressed and encodes strong priors for how generalization ought to occur. This results in a spectrum of policy structures, ranging from highly general (but often sample-inefficient) to highly constrained (but potentially more sample-efficient) representations.

\textbf{Nonparametric Policies:} 
The most expressive policy representations are nonparametric, growing as needed with task or data complexity. This category includes nearest neighbor-based approaches, Gaussian processes \cite{rasmussen2004gaussian}, Riemannian Motion Policies \cite{Ratliff2018}, and locally-weighted regression \cite{atkeson1997locally, vijayakumar2000locally}.  These representations are the most flexible and data-driven, but they also typically require large amounts of data to produce high-quality generalization.  This representational paradigm has been successful in manipulation tasks in which very little task knowledge is given \textit{a priori} to the robot, such as an 88-dimensional octopus arm control problem in which Gaussian Process Temporal Difference learning is used \cite{engel2006learning}, and 50-dimensional control of a SARCOS arm via locally weighted projection regression \cite{vijayakumar2000locally}.

\textbf{Generic Fixed-size Parametric Policies:} 
More commonly, fixed-complexity parametric policy representations are used, making stronger assumptions about the complexity and structure of the policy.  Common parametric policy representations include look-up tables \cite{sutton1998reinforcement}, linear combinations of basis functions such as tile coding, the Fourier basis \cite{konidaris2011value,Konidaris12a}, neural networks \cite{levine2016end}, decision tree classifiers \cite{quinlan1986induction, huang2002strategy}, and support vector machines \cite{cortes1995support, ross2011reduction}.  While any given choice of parameters has a fixed representational power, a great deal of engineering flexibility remains in choosing the number and definition of those parameters. Design decisions must also be made about which features will interact and which will contribute independently to the policy (e.g. the fully-coupled vs. independent Fourier basis \cite{konidaris2011value}), or more general decisions regarding inductive bias and representational power (e.g. neural network architectures).

The choice of a fixed parameterization makes particular assumptions about how generalization ought to occur across state-action pairs.  For example, tabular representations can represent any (discrete) function, but do not intrinsically support generalization to unseen states and actions.  By contrast, policies comprised of linear combinations of basis functions (for both discrete and continuous state spaces) more naturally generalize to novel situations since each parameter (a basis weight) affects the policy globally, assuming basis functions with global support.  However, the success of such generalization relies on the correctness of assumptions about the smoothness and ``shape" of the policy, as encoded by the choice of basis functions.  Some parametric forms make more specific structural assumptions---for example, by construction, convolutional neural networks \cite{krizhevsky2012imagenet,levine2016end,agrawal2016learning,mahler2017dex} support a degree of invariance to spatial translation of inputs.

\textbf{Restricted Parametric Policies:} 
The expressiveness of a policy is only limited by the underlying policy representation; however, sample complexity, overfitting, and generalization often get worse as representational power increases. This has lead to the development of specialized policy representations that have limited representational power, but exploit the structure of robotics problems to learn and generalize from less data. However, the quality of the generalization in such methods is highly dependent upon the accuracy of the underlying assumptions that are made. Such methods lie on a spectrum of how restrictive they are; for example, structured neural network architectures (e.g. value iteration networks \cite{tamar2016value}, schema networks \cite{kansky2017schema}, etc.) impose somewhat moderate restrictions, whereas other methods impose much stronger restrictions, such as forcing the policy to obey a particular set of parameterized differential equations \cite{schaal2006dynamic}.
However, it is not always clear how to choose the right point along the spectrum from a completely general to highly-specific policy class.  The following are some common examples of restricted policy classes.

Linear Quadratic Regulators (LQR) \cite{zhou1996robust} are commonly used to stabilize around (possibly learned) trajectories or points, in which the cost is assumed to be quadratic in state and the dynamics are linear in state.  These restrictions are relaxed in Iterative LQR \cite{fu2016one} and Differential Dynamic Programming \cite{tassa2014control,yamaguchi2015differential}, in which nonlinear dynamics and nonquadratic costs can be approximated as locally linear and locally quadratic, respectively. LQR-based controllers have also been chained together as LQR-trees, allowing coverage of a larger, more complex space \cite{tedrake2010lqr}.  LQR-like methods also assume that the state is fully observable; Linear Quadratic Gaussian (LQG) control methods further generalize LQR by using a Kalman filter for state estimation, in conjunction with an LQR controller \cite{levine2015learning,Platt2010}.  LQR/LQG controllers are optimal when the linear-quadratic assumptions are met, but require full knowledge of the dynamics and cost function, as well as a known trajectory around which to stabilize.  Thus, generalization in this case simply translates to being able to stabilize around this known trajectory optimally from anywhere in the state space.

Dynamic Movement Primitives (DMPs) support the learning of simple, generalizable policies from a small number of demonstrations, which can be further improved via reinforcement learning \cite{schaal2006dynamic,pastor2009learning}. DMPs leverage the fact that many robotic movements can be decomposed into two primary parts---a goal configuration in some reference frame and a ``shape" of the motion.  For example, screwing in a screw requires a turning motion to get to the desired final configuration with respect to the hole.  DMPs use a set of differential equations to implement a spring-mass-damper that provably drives the system to an adjustable goal from any starting location, while also including the influence of a nonlinear function that preserves the desired shape of the movement. DMP controllers can be learned from very little data, but typically only generalize well in cases in which a good solution policy broadly can be described by a single prototypical motion shape primitive.

Whereas DMPs generate deterministic policies, other approaches treat the problem probabilistically, allowing stochastic policies to be learned.  ProMPs are a straightforward probabilistic variant of DMPs that can produce distributions of trajectories \cite{paraschos2013probabilistic}.  Gaussian Mixture Regression (GMR) \cite{calinon2007learning} models the likelihood of states over time as a mixture of Gaussians, allowing for a multimodal distribution of trajectories that can encode multiple ways of performing a task, as well as the acceptable variance in different parts of the task.

\textbf{Goal-based Policies:}
At the far end of the spectrum, the most restrictive policy representations are primarily parameterized by a goal configuration. Given that the goal configuration is the primary parameter, these methods are typically either fixed strategies (such as splining to a goal point \cite{SU2016}, or between keyframes \cite{Akgun2012}) or have a very small number of adjustable parameters, such as a PID controller or motion planner.

\subsection{Reinforcement Learning}
     For any given policy representation, reinforcement learning (RL) \cite{sutton1998reinforcement} can be used to learn policy parameters for skill controllers. In robotics domains, tasks addressed with RL are usually \textit{episodic}, with a fixed number of time steps or a set of terminal states that end the episode (e.g. reaching a particular object configuration), but occasionally may be \textit{continuing} tasks with infinite horizons (e.g.  placing a continual stream of objects into bins as fast as possible).  
     
     There are many challenges in applying RL to robotics.  Due to the time it takes to collect data on physical robots, the tradeoff between exploration and exploitation becomes significantly more important, compared to problems for which fast, accurate simulators exist.  Furthermore, few problems in robotics can be strictly characterized as MDPs, but instead exhibit partial observability \cite{Platt2011} and nonstationarity \cite{padakandla2019reinforcement} (though it often remains practical to cast and solve these problems as MDPs, nonetheless). Since many tasks in robotics are multi-objective (e.g. pick up the mug, use as little energy as possible, and don't collide with anything), it can be difficult to define appropriate reward functions that elicit the desired behavior \cite{hadfield2016cooperative}. Due to the episodic nature of most robotics tasks, rewards tend to be sparse and difficult to learn from.  Robotics problems also often have high-dimensional continuous state features, as well as multi-dimensional continuous actions, making policy learning challenging.
	    
	Here, we categorize RL algorithms using three primary criteria: (1) model-based or model-free, (2) whether or not they compute a value function, and in what manner they use it, and (3) on-policy or off-policy.

\textbf{Model-Based RL:} 
Accurate models of transition dynamics are rarely available \textit{a priori} and it is often difficult to learn transition models from data. Nonetheless, there have been notable model-based successes in robotic manipulation in which an approximate model is learned from data \cite{deisenroth2011pilco,levine2015learning,lenz2015deepmpc,Kumar2016,FinnL17,schenck2018spnets}.  In these examples, the model is typically used to guide exploration and policy search, leading to greatly improved data efficiency.  More generally, the primary benefits of model-based RL in robotics are that (1) in some domains, an approximate model is simpler to learn than the optimal policy (since supervised learning is generally easier than RL) (2) models allow for re-planning / re-learning on-the-fly if the task changes, and (3) models allow for certain types of data collection that are difficult or impossible in the real world (such as resetting the world to an exact state and trying different actions to observe their outcome).  The primary disadvantage of model-based methods, aside from the difficulty of obtaining or learning models, is that incorrect models typically add bias into learning. However, some methods mitigate this by directly reasoning about uncertainty \cite{deisenroth2011pilco}, whereas methods such as doubly robust off-policy evaluation \cite{jiang2015doubly} use a model as a control variate for variance reduction, sidestepping the bias issue.
More details on model learning can be found in Section \ref{sec:modellearning}.

\textbf{Model-Free RL:} Model-free methods learn policies directly from experiences of the robot, without building or having access to a model.  
%The obvious advantages of the model-free approach is that there is no need for a model, and that under mild assumptions, most model-free methods are unbiased. 
When the dynamics of the environment are complex, as they often are in contact-rich manipulation tasks, it can be significantly easier to learn a good policy than a model with similar performance.  Model-free methods can learn how to implicitly take advantage of complex dynamics without actually modeling them—--for example, finding a motion that can pour water successfully without understanding fluid dynamics.  The downside of the model-free approaches is that they cannot easily adapt to new goals without additional experience in the world, unlike model-based approaches.  Furthermore, they typically require a large (sometimes prohibitive) amount of experience to learn.  For this reason, some recent approaches have sought to combine the benefits of model-based and model-free methods \cite{gu2016continuous, chebotar2017combining, pong2018temporal, feinberg2018model,Englert2016}.

\textbf{Value Function Methods:} These methods aim to learn the value of states (or more commonly, state-action pairs)---the expected return when following a particular policy, starting from that state(-action pair).  
%These are sometimes referred to as ``critic-only" methods, as the value function acts as a ``critic" that evaluates the goodness of a state(-action pair).  
Value functions methods are known to be low variance, highly sample efficient, and in discrete domains (or continuous domains with linear function approximation), many such methods can be proven to converge to globally optimal solutions.  However, value function methods are typically brittle in the face of noise or poor or underpowered state features, and do not scale well to high-dimensional state spaces, all of which are common in robot manipulation tasks. Furthermore, they are not compatible with continuous actions, severely limiting their application to robotics, except where discretization is acceptable. Actions are often discretized at higher levels of abstraction when they represent the execution of entire skills, rather than primitive actions \cite{Kroemer}.  Finally, it is difficult to build in useful structure and task knowledge when using value function methods, since the policy is represented implicitly. Nonetheless, value function-based methods such as Deep Q-Networks \cite{mnih2015human} and extensions thereof have been applied successfully to robotics tasks \cite {zhang2015towards,kalashnikov2018qt}.

\textbf{Policy Search Methods:} Policy search methods parameterize a policy directly and search for parameters that perform well, rather than deriving a policy implicitly from a value function. Policy search methods have become popular in robotics, as they are robust to noise and poor features and naturally handle continuous actions,
(the direct parameterization of the policy eliminates the need to perform a maximization over values of actions).  They also often scale well to high-dimensional state spaces since the difficulty of policy search is more directly related to the complexity of a good (or optimal) policy, rather than the size of the underlying state space.  

Pure policy search approaches eschew learning a value function entirely.  These are sometimes referred to as ``actor-only'' methods, since there is a direct parameterization of the policy, rather than a ``critic'' value function.  Actor-only approaches include the gradient-based method REINFORCE \cite{williams1992simple}, for use when the policy is differentiable with respect to its parameters.  However, REINFORCE tends to suffer from high variance due to noisy sample-based estimates of the policy gradient (though a variance-reducing baseline can be used to partially mitigate this) and is only locally optimal.
Actor-only policy search also includes gradient-free optimization methods \cite{de2005tutorial,mannor2003cross, hansen2001completely,davis1991handbook,theodorou2010generalized}, which are usable even when the policy is non-differentiable, and in some cases (e.g. genetic algorithms) are globally optimal, given sufficient search time.  Unsurprisingly, these advantages often come at the cost of reduced sample efficiency compared to gradient-based methods.

In contrast to actor-only approaches, actor-critic policy search methods \cite{sutton2000policy,peters2008natural,peters2010relative,lillicrap2015continuous,mnih2016asynchronous,schulman2015trust,schulman2017proximal,wang2016sample,wu2017scalable,haarnoja2018soft} use both a value function (the critic) and a directly parameterized policy (the actor), which often share parameters.  These methods typically have most of the advantages of actor-only methods (the ability to handle continuous actions, robustness to noise, scaling, etc), but use a bootstrapped value function to reduce the variance of gradient estimates, thereby gaining some of the sample efficiency and low-variance of critic-only methods.  
The unique advantages of actor-critic methods have made them state of the art in many robotic manipulation tasks, as well as in the larger field of deep reinforcement learning.

\textbf{On-Policy vs Off-Policy Learning:} 
When using RL to improve a policy, on-policy algorithms are restricted to using data collected from executions of that specific policy (e.g. SARSA \cite{sutton1998reinforcement}), whereas off-policy algorithms are able to use data gathered by any arbitrary policy in the same environment (e.g. Q-learning \cite{watkins1992q}). This distinction has several consequences which are notable in robotic domains for which data collection has significant costs.  On-policy algorithms are not able to re-use historical data during the learning process---every time the behavior policy is updated, all previously collected data becomes off-policy.  
It is also often desirable to use collected data to simultaneously learn policies for multiple skills, rather than only a single policy; such learning is impossible in an on-policy setting.  
However, despite these advantages, off-policy learning has the significant downside of being known to diverge under some conditions when used with function approximation, even when it is linear \cite{sutton1998reinforcement}. By contrast, on-policy methods are known to converge under mild assumptions with linear function approximation \cite{sutton1998reinforcement}.

\textbf{Exploration Strategies:}
A significant, but often overlooked, part of policy learning is the exploration strategy employed by the agent. This design decision can have enormous impact on the speed of learning, based on the order in which various policies are explored.
In physical robotic manipulation tasks, this is a particularly important choice, as data is much more difficult to collect than in simulated domains.  Furthermore, in the robotics setting, there are concerns related to safety and possible damage to the environment during exploration, as will be further discussed in Section \ref{sec:safety}.  

The most common exploration strategies involve adding some form of noise to action selection, whether it be in the form of direct perturbation of policy parameters \cite{theodorou2010generalized}, Gaussian noise added to continuous actions \cite{lillicrap2015continuous}, or epsilon-greedy or softmax action selection in discrete action spaces \cite{mnih2015human}. Unsurprisingly, random exploration often fails to efficiently explore policy space, since it is typically confined to a local region near the current policy and may evaluate many similar, redundant policies.

To address these shortcomings, exploration strategies based on the psychological concept of \textit{intrinsic motivation} \cite{Chentanez2005,oudeyer2009intrinsic} have used metrics such as novelty \cite{huang2002novelty,bellemare2016unifying, hart2009intrinsic,ecoffet2019go,burda2018exploration}, behavioral diversity \cite{lynch2019learning}, uncertainty \cite{martinez2007active}, and empowerment \cite{mohamed2015variational} to diversify exploration in a more effective way.  However, these methods are heuristic-based and may still lead to poor performance.  In fact, many essentially aim to use intrinsic motivation to uniformly explore the state space, ignoring structure that may provide clues about the relevance of different parts of the state space to the problem at hand.  By contrast, other approaches have taken advantage of structure specific to robotic manipulation by directing exploration to discover reusable object affordances \cite{montesano2008learning}, or learning exploration strategies that exploit the distribution of problems the agent might face via metalearning \cite{xu2018learning}.  Finally, as discussed in the next subsection, the difficulty of exploration in RL is sometimes overcome by leveraging demonstrations of good behavior, rather than learning from scratch.

\subsection{Imitation Learning}
In contrast to  reinforcement learning, which learns from a robot’s experiences in the world (or a model of it), imitation learning \cite{schaal1999imitation, argall2009survey} aims to learn about tasks from demonstration trajectories. This can be thought of as a form of programming, but one in which the user simply shows the robot what to do instead of writing code to describe the desired behavior. Learning from demonstration data has been extensively studied in several different settings, because it can enable the robot to leverage the existing task expertise of (potentially non-expert) humans to (1) bypass time-consuming exploration that would be required in a reinforcement learning setting, (2) communicate user preferences for how a task ought to be done, and (3) describe concepts, such as a good tennis swing, that may be difficult to specify formally or programmatically.
It is worth noting that imitation learning and reinforcement learning are not mutually exclusive; in fact, it is common for imitation learning to be followed by reinforcement learning for policy improvement \cite{kober2009policy,taylor2011integrating}.  

Demonstrations in imitation learning are typically represented as trajectories of states or state-action pairs. There are several mechanisms by which a robot may acquire such demonstration trajectories, including teleoperation, shadowing, kinesthetic teaching, motion capture, which are discussed in greater detail in the survey by Argall et. al \cite{argall2009survey}.  More recently, keyframe demonstrations \cite{Akgun2012}, virtual reality demonstrations \cite{zhang2018deep,yan2018learning}, and video demonstrations in the so-called learning from observation setting \cite{liu2017imitation} have also become more commonly used.
	    
\textbf{Behavioral Cloning:}
The simplest way to use demonstration data to
learn a motor skill is to use it as supervised training data to learn the robot's policy.  This is commonly called \textit{behavioral cloning}. Recall that a deterministic policy $\pi$ is a mapping from states to actions: $\pi: S \rightarrow A$, whereas a stochastic policy maps state-action pairs to probabilities (which sum to 1 at each state when marginalizing over actions): $\pi: S \times A \rightarrow \mathcal{R}$.  
The demonstration provides a set of state-action pairs $(s_i, a_i)$, that can be used as training data for a supervised learning algorithm to learn policy parameters that should, ideally, be able to reproduce the demonstrated behavior in novel scenarios.

Behavioral cloning is often used as a stand-alone learning method \cite{ross2011reduction,bagnell2007boosting,torabi2018behavioral,pastor2009learning,cederborg2010incremental,Akgun2012,duan2017one,hayes1994robot,schaal2005learning}, as well as a way to provide a better starting point for reinforcement learning \cite{pastor2011skill}, though this additionally requires the specification of a reward function for RL to optimize. Other recent work has focused on expanding the purview of behavioral cloning by unifying imitation learning and planning via probabilistic inference \cite{rana2017skill}, utilizing additional modalities such as haptic input \cite{kormushev2011imitation}, learning to recognize and recover from errors when imitating \cite{pastor2011online,pastor2011skill}, and scaling behavioral cloning to complex, multi-step tasks such as IKEA furniture assembly \cite{niekum2015learning}.

\textbf{Reward Inference:}
Rather than learning a policy directly from demonstration data, an alternative
approach is to attempt to infer the underlying reward function that the demonstrator was trying to optimize. 
This approach aims to extract the intent of the motion, rather than the low-level details of the motion itself. 
This approach is typically called inverse reinforcement learning (IRL) \cite{ng2000algorithms}, apprenticeship learning \cite{abbeel2004apprenticeship,Boularias2012}, or inverse optimal control \cite{moylan1971nonlinear,Englert2017,Englert2018}.
The inferred reward function can then be optimized via reinforcement learning to learn a policy for the task.

The IRL paradigm has several advantages. First, if the reward function is a function of the objects or features in the world and not the agent's kinematics, then it can be naturally ported from human to robot (or between different robots) without encountering the correspondence problem. In addition, reward functions are often sparse, thereby providing a natural means of generalizing from a small amount of training data, even in very large state spaces. In addition, the human's behavior may encode a great deal of background information about the task---for example, that an open can of soda should be kept upright when it is moved---that are easy to encode in the reward function but more complex to encode in a policy, and which can be reused in later contexts.
Unfortunately, IRL also presents several difficulties.  Most notably, the IRL problem is fundamentally ill-posed---infinitely many reward functions exist that result in the same optimal policy \cite{ng2000algorithms}.  Thus, the differentiation between many IRL algorithms lies in the metrics that they use to disambiguate or show preference for certain reward functions \cite{abbeel2004apprenticeship,abbeel2010autonomous,Boularias2012,aghasadeghi2011maximum,Englert2017,Englert2018}.

Maximum Entropy IRL \cite{ziebart2010modelingB} addresses the problems of demonstrator suboptimality and ill-posedness by leveraging a probabilistic framework and the the principle of maximum entropy to disambiguate possible reward functions.  Specifically, they develop an algorithm that assigns equal probability to all trajectories that would receive equal return under a given reward function and then use this distribution to take gradient steps toward reward functions that better match the feature counts of the demonstrations \cite{ziebart2008maximum}, while avoiding having any additional preferences other than those indicated by the data. 
Rather than generating a point estimate of a reward function, which forces an algorithm to face the ill-posedness of IRL head on, Bayesian IRL \cite{ramachandran2007bayesian}  instead uses Markov Chain Monte Carlo to sample from the distribution of all possible reward functions, given the demonstrations.  %Bayesian IRL can learn from partial demonstrations (or even single state-action pairs), rather than full demonstration trajectories.
Finally, in the more restricted case of linearly-solvable MDPs, the IRL problem is well-posed, avoiding these problems \cite{dvijotham2010inverse}.

All of the IRL algorithms mentioned so far rely on reward functions specified as a linear combination of features.  While this does not restrict the expressivity of reward functions in practice (more complex features can always be provided), it burdens the designer of the system to ensure that features can be learned from in a linear manner.  By contrast, Gaussian Process and nonparametric IRL \cite{levine2011nonlinear,Englert2017} and various neural network-based methods \cite{ho2016generative, finn2016guided, wulfmeier2015maximum,fu2017learning} aim to partly relieve this burden by searching for reward functions that are a nonlinear function of state features.  However, such flexibility in representation requires careful regularization to avoid overfitting \cite{finn2016guided}.  

Many of the aforementioned methods have an MDP solver in the inner loop of the algorithm.  Computational costs aside, this is especially problematic for robotics settings in which a model is not available and experience is expensive to collect.  Some recent IRL methods that have been shown to work in real robotic domains sidestep this obstacle by alternating reward optimization and policy optimization steps \cite{finn2016guided} or framing IRL as a more direct policy search problem that performs feature matching \cite{doerr2015direct,ho2016generative}.  If available, ranked demonstrations can be used to get rid of the need for inference-time policy optimization or MDP solving entirely, by converting the IRL problem to a purely supervised problem; furthermore, this approach allows the robot to potentially outperform the demonstrator \cite{brown2019trex,brown2019better}. Alternately, active learning techniques have been used to reduce the computational complexity of IRL \cite{brown2019risk,cui2018active,lopes2009active}, as well as strategies that make non-I.I.D. assumptions about the informativeness of the demonstrator \cite{brown2018machine,kamalaruban2019interactive}.  Finally, outside of an imitation learning framework, goals for the robot are sometimes specified via natural language commands that must be interpreted in the context of the scene \cite{tellex2011understanding}.

\textbf{Learning from Observation:} A relatively new area of inquiry aims to learn from demonstrations, even when no action labels are available and the state is not exactly known.  For example, a robot may visually observe a human performing a task, but only have access to raw pixel data and not the true underlying state of the world, nor the actions that the human took.  This problem is referred to Learning from Observation (LfO), and several recent approaches have addressed problems including unsupervised human-robot correspondence learning \cite{sermanet2017time}, context translation \cite{liu2017imitation}, adversarial behavioral cloning \cite{torabi2018behavioral}, and IRL from unsegmented multi-step video demonstrations \cite{goo2018learning,Yang2015}. In an extreme version of the LfO setting, the agents is expected to infer an objective from single-frame goal-state images, rather than a full trajectory of observations \cite{zeng2018semantic,xie2018few}.

\textbf{Corrective Interactions:} Rather than learning from full demonstrations in batch, it is often advantageous to solicit (potentially partial) corrective demonstrations or other forms of feedback over time. For example, a human could intervene in a pouring task and adapt the angle of the cup and the robot's hand mid-pour.   This provides a natural mechanism to collect data in situations where it is most needed---for example, situations in which mistakes are being made, or where the robot is highly unsure of what to do.  Some approaches actively ask users for additional (partial) demonstrations in areas of the state space in which confidence is low \cite{chernova2009interactive} or risk is high \cite{brown2019risk}, while others rely on a human user to identify when a mistake has been made \cite{niekum2015learning}.  Higher level information can also be used to make more robust corrections, such as grounded predicate-based annotations of corrections \cite{mueller2018robust} and action suggestions in a high-level finite state machine \cite{holtz2018interactive}. The robot can also actively solicit assistance when needed, for example, via natural language \cite{knepper2015recovering}.  Finally, rather than using corrective demonstrations, the TAMER framework uses real-time numeric human feedback about the robot's performance to correct and shape behavior \cite{knox2013training}.

\subsection{Skill Transfer}
\label{sec:transfer}
Given the high sample complexity of learning in complex robotics tasks, skills learned in one task are often transferred to other tasks via a variety of mechanisms, thereby increasing the efficiency of learning.  

\textbf{Direct Skill Re-use:}     
One of the simplest ways to transfer a skill policy is to directly re-use it in a new task related to the one it was learned on.  Typically, some amount of adaptation is required to achieve good performance in the new task.  One simple way to perform such refinement is to initialize a new skill with an existing skill's policy parameters and adapt them for the new task via reinforcement learning \cite{pastor2009learning}.  However, a naive realization of this approach only works when the original task and the new task have identical state representations.  When this is not the case, transfer can still occur via a subset of shared state features that retain semantics across tasks.  This is sometimes called an \textit{agent space} \cite{konidaris2007building}, since these are typically agent-centric, generic features (e.g laser scanner readings), rather than problem-specific features, or a \textit{deictic representation} \cite{Platt19}.  Since an agent space only covers some subset of the features in a problem, transfer occurs via initialization of a value function, rather than policy parameters.

More generally, it is often useful to find a state \textit{abstraction}---a minimal subset of state features required to perform some skill. Abstractions facilitate transfer by explicitly ignoring parts of the state that are irrelevant for a particular skill, which could otherwise serve as distractors (e.g. irrelevant objects), as well as allowing transfer to state spaces of different sizes (as also seen in agent spaces, which are a type of abstraction).  In some state spaces, such as the visual domain, the state space is not factored in a manner that makes abstraction easy (e.g. the texture of an object is not a separate state feature, but distributed across many pixels). One popular way of forcing a deep neural network to abstract away complex variables such as texture and color is \textit{domain randomization} \cite{tobin2017domain,bousmalis2017using}, which is discussed in greater detail later in this section. 
	            
\textbf{Parameterized Skills:}    
In certain task families, only some aspects of the task context change, while all other task semantics remain the same or are irrelevant.  For example, it may be desirable to transfer a policy that can hit one goal location on a dartboard, in order to hit a different goal location; similarly, transfer learning could acquire a policy for completing a pendulum swing-up task with different pendulum lengths and masses.  In these restricted cases, specialized \textit{parameterized skills} can be learned that facilitate transfer via mechanisms that modulate the policy based on the aspect of the task  context parameter that is changing \cite{Calinon2018,Englert2018a}.  

Dynamic Movement Primitives \cite{schaal2006dynamic,pastor2009learning} use a simple spring-mass-damper system to smoothly adjust to new initial and goal locations.  Another approach uses manifold learning to smoothly modulate policy parameters based on a variable task parameter \cite{da2014learning}.  Contextual policy search uses a hierarchical, two-level policy for low-level control and generalization across contexts, respectively \cite{kupcsik2013data}. Universal value function approximators \cite{schaul2015universal,andrychowicz2017hindsight} track a value function for all $\langle$state, action, goal$\rangle$ triplets, rather than only $\langle$state, action$\rangle$ pairs, allowing policy similarities across nearby goals to be leveraged explicitly.          
	            
\textbf{Metalearning:}   
Rather than re-using a skill directly for initialization, metalearning approaches aim to ``learn to learn"---in other words, learn something about a distribution of tasks that allows for more efficient learning on any particular task from that distribution in the future. Thus, metalearning facilitates transfer within a task family by performing learning across samples from a task distribution at training time, rather than performing transfer sequentially and on-line as problems are encountered, as in the direct re-use case. 

Model Agnostic Metalearning (MAML) \cite{finn2017model} searches for a set of policy parameters that can be adapted quickly for particular tasks drawn from some distribution. Reptile \cite{nichol2018reptile} simplifies MAML by replacing a complex optimization with an approximate approach that only requires standard stochastic gradient descent to be performed on each sampled task individually.  Another related metalearning approach learns an attention-based strategy that allows the robot to imitate novel manipulation tasks, such as block-stacking, from a single demonstration \cite{duan2017one}.

Other forms of metalearning have focused on reward functions rather than policies.  While many problems in robotic manipulation have simple sparse reward formulations (e.g. +1 when peg is in hole, -1 otherwise), \textit{potential-based} shaping rewards \cite{ng1999policy} can be added to any reward function to encourage faster learning (across some distribution of problems) without changing the optimal policy of the MDPs.  More generally, modified reward functions can help to overcome multiple forms of \textit{agent boundedness} \cite{sorg2010internal} and can be found via gradient descent \cite{sorg2010reward}, genetic programming \cite{niekum2010genetic}, or other evolutionary methods \cite{houthooft2018evolved}.  

Some metalearning approaches have been developed to directly modify the representation of the policy itself, or other parts of the learning algorithm, rather than only the settings of policy or reward parameters.  This includes methods that evolve the structures of neural networks \cite{stanley2009hypercube}, or that learn skill embeddings \cite{hausman2018learning}, structured exploration strategies \cite{gupta2018meta}, transfer feature relevances between families \cite{KroemerMeta},reusable network modules \cite{alet2018modular}, or that co-learn a differentiable model and trajectory optimizer \cite{srinivas2018universal}.

\textbf{Domain Adaptation:}
In contrast to parameterized skills, some task families retain all of their high-level semantics across instances, differing only in lower-level details.  In these cases, \textit{domain adaptation} techniques are commonly used to bridge the so-called domain gap between two (or more) domains.  For example, in the ``sim2real" problem, when switching from a simulated task to a physical robot, the low-level statistics of the visual scene and physics may change, while the high-level steps and goals of the task stay fixed. Grounded action transformations address the physics domain gap by iteratively modifying agent-simulator interactions to better match real-world data, even when the simulator is a fixed black box \cite{hanna2017grounded}.  Domain randomization is a data synthesis technique used to address visual domain gaps by randomizing environmental attributes such as texture, color, and dynamics in simulation, as to force the system to be invariant to statistical changes in the properties \cite{tobin2017domain,bousmalis2017using,peng2017sim}.  A different type of visual shift can result from changes in viewpoint, often encountered in the setting of learning from unstructured videos. Approaches to this problem have included unsupervised learning of correspondences between first-person and third-person task demonstrations \cite{stadie2017third}, as well as imitation-from-observation approaches that work from multi-viewpoint data \cite{liu2017imitation}. Other approaches attempt to learn disentangled representations that lead naturally to robust polices that are tolerant of domain shift \cite{higgins2017darla}.

In some tasks, the state and action space may change in a manner that requires an analogy to be made between the domains, as in the case of transferring a manipulation policy between two robot arms with different kinematic structures by finding a shared latent space between the two policy representations \cite{gupta2017learning}.  Time Contrastive Networks take a self-supervised metric-learning approach to a similar problem, learning representations of behaviors that are invariant to various visual factors, enabling robot imitation of human behaviors without an explicit correspondence \cite{sermanet2017time}.  Other methods perform a more direct mapping between states and actions \cite{taylor2007trans}, but assume that there exists a complete mapping, while the aforementioned latent space approach is able to discover partial mappings.

\textbf{Sequential Transfer and Curriculum Learning:}
While the aforementioned transfer paradigms primarily consider instances of transfer individually, it is sometimes advantageous to view multiple instances transfer as a sequential learning problem.  For example, Progressive Neural Networks \cite{rusu2016progressive} use a neural network to learn an initial task, transferring this knowledge to subsequent tasks by freezing the learned weights, adding a new column of layers for the new task, and making lateral layer-wise connections to the previous task's neurons.  
Sequential transfer learning is also a useful paradigm for mastering tasks that are too difficult to learn from scratch. Rather than attack the final task directly, \textit{curriculum learning} presents a sequence of tasks of increasing difficulty to the agent, which provides a smoother gradient for learning and can make positive rewards significantly less sparse in an RL setting.  Such methods often use curricula provided by human designers \cite{bengio2009curriculum,sanger1994neural,pinto2016supersizing}, though several recent methods seek to automatically generate a curriculum \cite{narvekar2016source,svetlik2017automatic,florensa2017reverse}.  

Scheduled Auxiliary Control \cite{riedmiller2018learning} learns to sequentially choose from a set of pre-defined auxiliary tasks in a manner that encourages efficient exploration and learning of the target task.  Guided Policy Search \cite{levine2015learning} first learns simple time-dependent controllers for individual task instances, which are then used to generate data to train a more complex neural network. Universal Value Function Approximators (UVFAs) \cite{schaul2015universal} can learn and transfer knowledge more efficiently in curriculum-like settings by estimating the value of a state and action \textit{conditioned upon the current goal}, rather than for a single fixed goal.  Hindsight Experience Replay (HER) provides additional goal settings for UVFAs by simply executing actions, and in hindsight imagining that every state that the agent ends up in was actually a goal \cite{andrychowicz2017hindsight}.

Reverse curriculum learning \cite{florensa2017reverse} begins at the goal and works backwards, trying to learn to get to the goal successfully from more and more distant states. This stands in contrast to standard curriculum learning that typically starts in some distant state, slowly moving the goal of the subtasks closer and closer to the true goal.  This idea is also related to that of skill chaining \cite{konidaris2009skill,Konidaris12a}, which will be discussed in further detail in Section \ref{sec:gentask}.

\subsection{Safety and Performance Guarantees}
\label{sec:safety}

Whether a policy is learned directly for a specific task or transferred from a previous task, an important but understudied question is ``How well will the policy perform across the distribution of situations that it will face?".  This is an especially important question in robotic manipulation, in which many future applications will require behaviors that are safe and correct with high confidence: robots that operate alongside humans in homes and workplaces must not cause injuries, destroy property, or damage themselves; safety-critical tasks such as surgery \cite{van2010superhuman} and nuclear waste disposal \cite{kim2002robotic} must be completed with a high degree of reliability; robots that work with populations that rely on them, such as the disabled or elderly \cite{goil2013using}, must be dependable. 
While the term ``safety" has taken on many different meanings in the literature, in this section, we take a broad view of safe learning to include all methods that aim to bound a performance metric with high confidence.  

\textbf{Performance Metrics:}  Most performance metrics for robotic manipulation tasks can be represented as cumulative reward under some reward function---for example, the preferences of completing a task accurately, finishing within a certain amount of time, or never (or with low probability) visiting certain states can all be described with particular reward functions and acceptable thresholds on cumulative reward, or \textit{return}.  Thus, for simplicity, we will discuss the problem of bounding performance as being equivalent to that of lower-bounding some function of return.  However, it is worth noting that certain preferences, such as some temporal logic statements, cannot be expressed as a reward function without augmenting the state space; some cannot be expressed by any reward function at all \cite{littman2017environment}.

The principle axis of variation regarding performance metrics is whether the \textit{expected} return of a policy is being bounded, or whether a risk-aware function of return is used---for example, a robust worst-case analysis \cite{ghavamzadeh2016safe}, a PAC-style bound \cite{strehl2006pac}, or bounded value-at-risk \cite{chow2015risk,brown2020safe}.  Risk-aware metrics are generally more appropriate for safety-critical tasks in which the robot could cause damage or injury in a single trial that goes poorly, whereas expected performance is often used in scenarios in which long-term performance (e.g. percentage of correctly manufactured items) matters more than the outcome of any single trial.

Finally, one important distinction to make is \textit{when} a policy must obey performance bounds. Some robotics tasks, such as learning to manipulate a fragile object, demand performance guarantees during the exploration and learning phase, rather than only after deployment of the final policy.  For example, in one recent work, demonstrations were used to constrain exploration in safety-critical manipulation tasks \cite{thananjeyan2019extending}. By contrast, other tasks  may lend themselves to safe practice, such that only the final policy needs to be accompanied by guarantees.  For example, a legged robot may learn a walking gait while attached to a support rig so that it cannot catastrophically fall while learning.

\textbf{Classes of Guarantees and Bounding Methods:} Given that standard policy learning in robotics is a challenging open problem, in part due to limited real-world data collection abilities, it is not surprising that safe learning is even more difficult. Safe learning methods are typically significantly more data-hungry and/or require more accurate models than standard learning methods, as they need to provide guarantees about the quality of the policy.  For this reason, very few safe learning methods (e.g. from the reinforcement learning community) have been used in robotics applications.  This is a significant gap in the current literature and an opportunity for future work that can provide strong performance guarantees in low-data and poor-model robotics settings.  

Safe learning methods fall roughly into two categories: Exact methods and probabilistic high-confidence bounds. Formal verification-based approaches are exact methods that use a known model, along with a safety specification (e.g. a finite-state automata) to guarantee (or bound the probability, in the stochastic case) that following a policy will not lead to an unsafe outcome \cite{fu2014probably,fainekos2009temporal,chinchali2012towards,chen2013synthesis,alshiekh2018safe}.  However, some research has focused on control synthesis that obeys linear temporal logic constraints without access to a model \cite{sadigh2014learning}. By contrast, probabilistically safe approaches aim to provide lower-bounds on expected return, rather than utilizing logical specifications of safety \cite{thomas2015high,thomas2015imp,jiang2015doubly,hanna2017bootstrapping,thomas2016data}. While probabilistic methods appear to be promising for robotics applications (since they do not require a model), to the best of our knowledge, they have not been used in real robotics problems, potentially due to their high sample complexity.

Thus far, we have only considered performance guarantees in a standard MDP setting, in which either (1) the reward function is known, or (2) samples from the reward function are available.  However, this description of the problem does not cover a common scenario that occurs when learning from demonstrations---having access to states and actions, but not rewards.  In the inverse reinforcement learning setting, several approaches have examined how to bound the policy loss between the the robot's policy (commonly a policy learning via RL under the inferred reward function) and the optimal policy under the ground-truth reward function of the demonstrator, even though it is unknown \cite{abbeel2004apprenticeship, brown2017efficient, brown2020safe}.

\section{Characterizing Skills by Preconditions and Effects}
\label{sec:prepost}
\label{sec:characterizing}

 Executing a manipulation skills alters the state of the robot's environment; if the robot is to use its skills to achieve a specific goal, it requires a model of those outcomes. This model is termed a \textit{postcondition}, and describes the regions in the state space that the robot will find itself in after skill execution. The robot must also model the circumstances under which the skill can be executed---termed its \textit{preconditions}.   Knowledge of these two models for each skill can be used to determine whether, for example, a sequence of skills can be executed from a particular state \cite{konidaris2018skills}. Pre- and postconditions are used to sequence skills (or actions) for task planning. The planner searches through sequences of actions with the constraint that the postconditions of each skill must fulfill the preconditions of the next skill.
 
 \subsection{Pre- and Postconditions as Propositions and Predicates}
 The pre- and postconditions of manipulation skills are typically represented abstractly as either \textit{propositions} or \textit{predicates} \cite{KRUGER2011740,Kulick2013,konidaris2018skills,BeetzCRAM,SHAPIRO2003,Ugur2015} that are either true or false in any specific state. For example, the robot may represent the outcome of a particular navigation skill using the boolean proposition \texttt{AtTableB}. It could also use a predicate representation, \texttt{At(TableB)}, which supports more compact and efficient representations, and allows for easier generalization across objects.  We  therefore use predicate-based symbols for the remainder of this section although most of the explanation also applies to proposition symbols.
		
		The \emph{grounding} or anchoring of a predicate  refers to the mapping between the (often continuous) low-level state and context and the predicate's truth value; it defines the meaning of the symbol \cite{konidaris2018skills,CoradeschiLW13a}. To avoid confusion, we will not use the term grounding for assigning objects to predicates as is done in the planning literature \cite{Russell2003}. 

        \textbf{Classifier Representation: }
		The grounding of a predicate can be modelled as a binary classifier \cite{konidaris2018skills,Kroemer2016,KonidarisIJCAI15}. The classifier represents the mapping from the state and context to either true or false. If a predicate is defined for a subset of objects, then the state and context features of those objects are used as the input to the predicate classifier \cite{Kroemer2016}. For example, the predicate \texttt{Grasped(KnifeA, RHand)} would consider the state and context of the knife and hand to determine if the knife was being grasped by the the hand. The robot may learn a probabilistic classifier to determine the likelihood that the predicate is true given the current state and context. In this manner, the robot may handle situations where the predicate is only sometimes valid, or when the classifier has been learned and the robot is uncertain of its true value in states it has not encountered before \cite{konidaris2018skills}.  

        \textbf{Distribution Representation: }
        Alternatively, predicates can be modelled as probability distributions over the state space \cite{konidaris2018skills,Detry2011}. The distributions are defined in the state and context space. For a non-probabilistic approach, the distribution defines the set of states and contexts where the distribution is true, and the predicate is false otherwise. For a probabilistic approach, the distribution defines a probability density over the states and contexts given that the predicate is true \cite{konidaris2018skills,Detry2011}. The distribution may be defined only for the objects assigned to the predicate. This distribution is useful for sampling states and contexts in which the predicate is true. 

        \textbf{Modularity and Transfer: }
        One precondition proposition and one postcondition proposition for each skill are sufficient for skill sequencing. These predicates can be monolithic representations that define the sets of states and contexts for that specific skill. However, by decomposing the conditions into modular predicates, the robot can share knowledge between different skills and tasks. These predicates can often be defined for subsets of objects, e.g., \texttt{Full(mug, water)} and \texttt{Grasped(mug, hand)}.  The predicates often define labels for individual objects, e.g., \texttt{Container(obj1)}, and the relationships between the objects, e.g., \texttt{Above(obj1,obj2)}. Modular predicates will often capture the contact-based manipulation modes between pairs of objects, e.g., \texttt{Grasped(Obj1,Hand)} or \texttt{On(Obj2,Obj3)}. Previous works have explored methods for learning specific predicates or discovering suitable sets of predicates \cite{Kulick2013,montesano2008learning,konidaris2018skills,Hjelm2014}. 
        
        Learning the preconditions and postconditions is generally easier when reusing predicates from previous tasks rather than learning  from scratch. However, some discrepancies may exist between tasks and thus require additional learning \cite{Ugur2015b}. For example, a robot may require a specific type of grasp for performing a task. The predicate \texttt{Grasped(obj3,hand)} is thus not sufficient. The robot should instead learn a general predicate for transferring knowledge between tasks, and a task-specific predicate for incorporating additional constraints \cite{Detry2017,tenPas2017,gualtieri2018,lu2019,Bohg,Hjelm2014}. The latter predicate is generally easier to learn given the former predicate. For example, a robot may learn to identify stable grasps for a wide range of objects, and subsequently learn to identify a subset of grasps for specific objects or to grasp handles of cooking utensils without relearning to grasp from scratch \cite{Detry2017,gualtieri2018,tenPas2017}.

\subsection{Learning Pre- and Postcondition Groundings}
        The robot can train the pre- and postcondition predicate classifiers using samples of states and contexts where the conditions were known to be true or false.  The ground truth labels could be provided by a human supervisor, although this approach would limit the autonomy of the robot and may require substantial expertise from the user. In some cases a human can  provide data for \emph{desired} post- and preconditions---what the skill should achieve, from which conditions---but the
        actual conditions will ultimately depend on the robot's capabilities.

        Instead of relying on manual labeling, the robot can learn the condition labels from experience. The labels for the preconditions can be learned given a fixed postcondition --- all states from which a skill execution leads to a state satisfying the postcondition are positive examples, and all other states are negative examples. The robot should use a probabilistic classifier to capture the stochasticity of the transitions.  The robot can thus obtain the precondition labels by executing the skill from different states and observing the resulting outcomes. 
        
        The postconditions are less trivial to define. A human may predefine a desired postcondition, but to achieve autonomy the robot must \emph{discover} different postconditions on its own. Distinct postconditions can be learned by clustering the outcomes of the skill from different initial states and contexts  \cite{Dogar2007,UgurSO09,Ugur2015}. Manipulation skills often have distinct effects, e.g., a box remained upright or toppled over, which can be extracted through clustering or by detecting salient events. Each distinct postcondition cluster will then be associated with its own preconditions.  This approach is often employed in developmental learning to discover specific useful skills from more general skill policies, e.g., grasping and pushing from reaching \cite{Oztop2004,Juett2018}.
        
        A more goal-oriented approach to specifying postconditions is to construct skills where the postcondition is  either the goal of the task or another skill's precondition \cite{konidaris2009skill,Konidaris11b}.   This approach directly learns pre- and postconditions for constructing  skills that can be sequentially executed, and may avoid learning conditions that are irrelevant to the robot's task set, but it introduces dependencies between the robot's skills.    
        
        Finally, pre- and postcondition predicates can be grounded in sensory data with the assistance of natural language, for example, as part of an interactive dialogue between a human and a robot \cite{thomason2016learning}. Conversely, knowledge of pre- and postconditions of skills can help to ground natural language commands to those skills, especially when the language description is incomplete \cite{misra2014tell}. For example, the command ``Stir the soup" may imply picking up a spoon first, which could be determined via the precondition to a stirring skill.

\subsection{Skill Monitoring and Outcome Detection}
        Most skills will have distinct pre- and postconditions with some of the predicates changing as a result of the skill execution, e.g., executing a grasping skill on a book should result in \texttt{Grasped(Book,RHand)=True} once the book has been grasped  \cite{DangGraspStability,garrettIJRR2017}. The predicates may also change due to errors in the skill execution. For example, when executing a placing skill, the predicate values \texttt{On(Book, Table)=False} or \texttt{InCollision(Book, Obstacle)=True}  would correspond to errors. To perform manipulation tasks robustly, the robot must monitor its skill executions and determine if and when it has achieved the intended outcome or whether an error has occurred.

\textbf{Learning Goal and Error Classifiers: }
        Goals and errors in skill executions can be modeled as distinct predicates with values that are switched for the postconditions. Detecting goals and errors can thus be modelled as a classification problem \cite{Bekiroglu2013,Rodriguez2010}.  Rather than using only the current state, the robot can incorporate action and sensor information from the entire skill execution \cite{Bekiroglu2013,madry2014,Rodriguez2010}, although it is often better to stop a skill early when an error occurs \cite{Pastor2012,SU2016,Sukhoy2012}. Transient events, such as vibrations from mode transitions or incipient slip can then be used to better detect the predicate switches \cite{Park2016,VeiHofPetHer2015,SU2016}. The robot can use a variety of sensor modalities, including vision and audio, to detect the predicate switches, e.g., a robot can use tactile sensing to determine if a grasp attempt succeeded \texttt{Grasped(obj)=True} \cite{Calandra2018,DangGraspStability,madry2014,Dang2013}.
        A robot can also learn optimal locations for placing a camera, or other sensor, to reliably verify if a desired postcondition has been fulfilled \cite{saran2017viewpoint}.

\textbf{Detecting Deviations from Nominal Sensory Values:}
        Stereotypical executions of a manipulation skill will usually result in similar sensations during the execution. Larger deviations from these nominal values often correspond to errors \cite{Yamaguchi2016}. Hidden Markov models can be used to track the successful progress of a skill's execution based on sensory signals \cite{Park2016,Lello2013,Hovland,Bekiroglu2010}. 
        The robot may also learn the nominal sensory signals as a regression problem \cite{pastor2011skill,Kappler,Sukhoy2012}. Significant deviations from the expected sensory values would then trigger the stopping of the current skill. An error may also trigger a corresponding recovery action \cite{Dang2013,Veiga2018,Yamaguchi2017}. Unlike the outcome classifiers from the previous section, these models are trained using only data from successful trials. 
        
\textbf{Verifying Predicates:} 
        The pre- and postconditions of skills often change the values of predicates corresponding to modes and constraints between objects. For example, a placing skill can make \texttt{On(DishC,DishB)=True} and an unlocking skill can make \texttt{Locked(DoorA)=False}. The robot can verify these swiches in the predicate values using \emph{interactive perception}. In some cases the predicate can be verified by directly performing the next skill in the sequence, e.g., attempting to lift an object after it has been grasped \cite{pinto2016supersizing,kalashnikov2018qt}. For contact constraints, the robot can often perform small perturbations to verify the predicate's final value \cite{Debus2004,Wang2019a}. In both examples, the robot can obtain a better estimate of the predicate by performing the additional skill and observing the effect.
        However, the additional skills can sometimes also change the predicates, and hence some care needs to be taken when verfiying predicates.

  \subsection{Predicates and Skill Synthesis}
 
Skills will often have additional high-level arguments that define how to execute the skill \cite{UgurOS11}. For example, a grasping skill may take in a grasping pose, or a scrubbing skill may take in a desired force \cite{Bohg2014}. In this manner, a higher-level policy may adapt the skill to its specific needs.
 
\textbf{Representing and Synthesizing Skill Parameters:}
        Similar to the predicate representations, these policy parameter arguments can be modeled as additional input features for the precondition classifier or as distributions over valid and invalid argument values \cite{Detry2011,JiangLearnToPlace}. Many of these arguments can even be thought of as virtual or desired object states \cite{JiangLearnToPlace}.  For example, one could sample potential hand positions for grasping objects and model each one as a symbol for specifying the grasping action \cite{garrettIJRR2017,Bohg2014}. To obtain valid grasp frames, the robot can learn a classifier to determine if sampled grasp frames will lead to successful grasps \cite{Saxena2008,Nguyen2014,Herzog2014,pinto2016supersizing}, or sample from a learned probability density over successful grasps or cached grasps from similar objects \cite{Detry2011,Brook2011,Goldfeder,Choi18}. The process of learning the pre- and post- conditions with additional policy arguments is thus similar to learning the preconditions without the arguments.
        
        Once the robot has learned the pre- and postconditions over the arguments, it can use these to select skill parameters for new situations. This process usually involves sampling different parameter values and evaluating them in terms of the pre- and postconditions. The robot could use sampling and optimization methods to select argument values with high likelihoods of successful skill executions. As these parameters are often lower dimensional than full skills, active learning and multi-armed bandit methods can be used to select suitable argument values \cite{Montesano2012,mahler2016dex,Kroemer2010}. When the positive distribution is learned directly, the robot can sample from the distribution and then evaluate if it is consistent with other predicates, e.g., not in collision \cite{Ciocarlie2014}. The arguments are usually selected in order to achieve certain predicates in the postconditions. One can therefore think of the argument selection process as \emph{predicate synthesis} \cite{AmesTK18}.

	    \textbf{Preconditions and Affordances: }
	    Affordances are an important concept in manipulation and a considerable amount of research has explored learning affordances for robots \cite{SahinAffordances,MinAffordances,JamoneAffordances}. The affordances of an object are the actions or manipulations that the object affords an agent \cite{gibson2014ecological}. An object that can be used to perform an action is thus said to \emph{afford} that action to the agent. For example, a ball affords bouncing, grasping, and rolling. Affordances are thus closely related to the preconditions of skills. As affordances connect objects to skills \cite{montesano2008learning}, affordance representations often include skill arguments that define how the skill would be executed\cite{UgurOS11,Kroemer2012}. 
	    
	    The exact usage of the term \emph{affordances} tends to vary across research papers \cite{JamoneAffordances}. In most cases, affordances can be seen as a form of partial preconditions. The affordances often correspond to specific predicates that the robot learns in the same manner as the preconditions. The other components of the precondition are usually constant or at least all valid, such that the success or failure of the skill only depends on the component being learned. As partial preconditions, some affordances are more specific than others, e.g., balls can be rolled versus balls on a plane within reach of the robot can be rolled. Ultimately, the full preconditions need to be fulfilled to perform the skill. However, the partial preconditions of affordances provide modularity and can help the robot to search for suitable objects for performing tasks. 

\section{Learning Compositional and Hierarchical Task Structures}
\label{sec:hierarchy}
In the previous sections, we have focused on learning models of individual objects, or on learning
to  perform or characterize individual motor skills. However, manipulation tasks often have a substantial \textit{modular} structure that can be exploited to improve performance across a family of tasks. Therefore, some research has attempted to decompose the solution to a manipulation task into \textit{component skills}.
Decomposing tasks this way has several advantages. Individual component skills can be learned more efficiently because each skill is shorter-horizon, resulting in a substantially easier learning problem and aiding exploration. Each skill can use its own internal skill-specific abstraction \cite{Diuk09,konidaris09skillspecificabstractions,vanSeijen13,Cobo14,Jiang15} that allows it to focus on only relevant objects and state features, decomposing a problem that may be high-dimensional if treated monolithically into one that is a sequence of low-dimensional subtasks. A skill's recurrence in different settings results in more opportunities to obtain relevant data, often offering the opportunity to generalize; conversely, reusing skills in multiple problem settings can avoid the need to relearn elements of the problem from scratch each time, resulting in faster per-task learning. Finally,  
these component skills create a \textit{hierarchical structure} that offers the opportunity to solve manipulation tasks using higher-level states and actions---resulting in an easier learning problem---than those in which the task was originally defined. 

\subsection{The Form of a  Motor Skill}

The core of hierarchical structure in manipulation learning tasks is identifying the component skills from which a solution can likely be assembled.  

Recall that skills are often modeled as options, each of which is described by a tuple $o = \left(I_o, \beta_o, \pi_o\right)$,
where:
\begin{itemize} 
\item $I_o: S \rightarrow \{0, 1\}$ is the initiation set, which corresponds
to a precondition as discussed in Section \ref{sec:characterizing}. 
\item $\beta_o: S \rightarrow \left[0, 1\right]$, the termination condition, describes the probability that option $o$ ceases execution upon reaching state $s$. This corresponds to a goal as discussed in Section \ref{sec:characterizing}, but is distinct from an effect; the goal is the set of states in which the skill \textit{could} (or perhaps \textit{should}) terminate, whereas the effect describes where it \textit{actually} terminates (typically either a subset of the goal or a distribution over states in the goal).
\item $\pi_o$ is the option policy. 
\end{itemize}
In many cases, option policies are defined indirectly using a reward function $R_o$, often consisting of a completion reward for reaching $\beta_o$ plus a background cost function. $\pi_o$ can then be obtained using any reinforcement learning algorithm, by treating $\beta_o$ as an absorbing goal set. The core question for finding component skills when solving a manipulation problem is therefore to define the relevant termination goal $\beta_o$---i.e, identify the target goal---from which $R_o$ can be constructed.

\label{sec:gentask}

The robot may construct a \textit{skill library}, consisting of a collection of multiple skills that can be frequently reused across tasks. This requires extracting a collection of skills, either from demonstrations, or from behaviors generated autonomously by the robot itself. 
The key question here is how to identify the skills, which is a difficult, and somewhat under-specified, challenge.
There are two dominant approaches in the literature: segmenting task solution trajectories into individual component skills, or directly including skill specification as part of the overall problem when learning to solve tasks. 

\subsection{Segmenting Trajectories into Component Skills}

One approach to identifying a skill library is to obtain solution trajectories
for a collection of tasks, and segment those trajectories into a
collection of skills that retroactively decompose the input trajectories. 
This is commonly done using demonstration trajectories, though it could also be performed on trajectories generated autonomously by the robot, typically after learning \cite{Hart08,Konidaris11b,Riano12}. 
However they are generated, the resulting trajectories must be segmented into component skills.
The literature contains a large array of methods for performing the segmentation, which we group into two broad categories.

\textbf{Segmentation Based on Skill Similarity:} The most direct approach is to segment demonstration trajectories into 
repeated subtasks, each of which may occur in many different contexts \cite{Dang2010,Meier2011}. Identifying such repeated tasks both reduces the size of the skill library (which in turn reduces the complexity of planning or learning which skill to use when) and maximizes the data available to learn each skill \cite{Lioutikov}. This requires a measure of \textit{skill similarity} that expresses a distance metric between two candidate skill segments, or more directly models the probability with which they were generated by the same skill. 
   Therefore, several approaches have used a measure of skill similarity to segment demonstration trajectories, often based on a variant of a Hidden Markov model, where 
   the demonstrated behavior is modeled as the result of the execution of a sequence of latent skills; segmentation in this case amounts to inferring the most likely sequence of skills. 
    
 The most direct approach \cite{Jenkins04,Chiappa10,Grollman10,niekum2015learning,Meier2011,Daniel16MLJ} 
 is \textit{policy similarity}, which 
    measures skill similarity by fitting the data to a parameterized policy class and  measuring distance in parameter space. Alternative approaches to policy 
    similarity fit models to the underlying value function \cite{Konidaris12a} or the unobserved reward function that the skill behavior is implicitly maximizing \cite{Ranchod15,Krishnan19,Michini12,Choi12,Babes11}. Some approaches segment demonstration trajectories and then merge similar skills in a separate post-processing step \cite{Konidaris12a}, while the most principled probabilistic approaches infer shared skills
    across a collection of trajectories as part of the segmentation process \cite{niekum2015learning,Ranchod15}. 
    % \cite{Lioutikov} where do they go?
Rather than using a parametric model, a latent space can be discovered that efficiently encodes skills, which are either learned from experience in a reinforcement learning context, or via a latent-space segmentation of trajectories in an imitation learning setting \cite{hausman2017multi,hausman2018learning}. Other recent approaches have relied on observations only, learning to segment videos of multi-step tasks into composable primitives \cite{goo2019learning,yu2018one,huang2018neural,xu2018neural}.

 A less direct approach is to measure skill-similarity based on \textit{pre- and
 post-condition similarity}, where the skill policy or trajectory itself is not 
used for segmentation. Instead, trajectory segments that achieve the same goal 
can be clustered together, while those that do so from very different initial conditions could be split \cite{Kroemer2014a}.  Thus, reaching and pushing motions are different skills due to their distinct pre- and post- conditions, i.e., moving or not moving an object, even if the skill policies are similar or the same.
 This approach is often used in developmental learning approaches to discover skills \cite{weng2001autonomous};  the robot executes a skill policy in a variety of scenarios and then clusters together the post- and pre- conditions to create distinct skills \cite{xie2018few,niekum2011clustering,UgurOS11,Ugur2015,Ugur2015b}. 

\textbf{Segmentation Based on Specific Events:} Several approaches
use pre-designated events to indicate skill boundaries. These can range from
hand-specified task-specific events to more generally applicable principles.

One common approach is to segment by \textit{salient sensory events} defined by 
haptic, tactile, audio, or visual cues \cite{Juett2018,SU2016,Fitzpatrick2006,Aksoy2011}. For example, insertion skills are easier to monitor if they result in a distinct \emph{click} upon successful completion. Similarly, we can determine if a light switch was properly pushed if the light goes on or the button cannot be pushed any further. Human-made objects are often designed to provide salient event feedback to reduce errors while using the objects. Such skills have the advantage that their termination conditions are easy to detect and monitor.

Another important class of pre-defined segmentation events is \textit{transitioning between modes} \cite{Baisero2015,Kroemer}.
        Switching between modes allows the robot to switch between the ability to interact  with different objects. Hence,  the robot must first transition to a suitable mode to manipulate an object. Skills for transitioning to specific modes allow the robot to decouple accessing a mode and using the mode to perform a manipulation task. Grasping, lifting, placing, and releasing are all examples of skills used to switch between modes in pick-and-place tasks. Mode transitions can also be verified by applying additional actions to determine if the skill execution was successful. However, multiple skills within a mode, e.g., squeezing, shaking, and tilting a held object, cannot be detected using only this type of decomposition.

\subsection{Discovering Skills While Solving Tasks}

An alternative approach is to discover component motor skills \textit{during}
the process of learning to solve one or more manipulation tasks. This has two
crucial advantages over solution trajectory segmentation. First, learned
skills could aid in solving the tasks to begin with; many complex manipulation
tasks cannot be solved directly without decomposing them into simpler tasks, so
it will be infeasible to require that a complete solution precede skill discovery. 
Second, imposing hierarchical structure \textit{during} learning can inject bias that results in much
more compact skill libraries. For example, there may be multiple ways to turn a switch; there is no reason to expect that a robot that independently learns to solve several tasks involving switches will find a similar policy each time. However, if the learned policy for turning a switch is identified and retained in the first task, then it will likely be reused in the second task. 

Broadly speaking, there are substantially fewer successful approaches that learn skills while solving tasks than there are approaches that retroactively segment,
in part because the problem is fundamentally harder than segmentation. For example, skill similarity approaches are difficult to apply here. However, some researchers have 
applied methods based on specific salient events \cite{Hart08} to trigger skill creation. These methods could be combined with skill chaining \cite{konidaris2009skill,Konidaris11b}---an approach where skills are constructed to either reach a salient event or to reach another skill's preconditions---to learn the motor skills online. 
An alternative is to structure the policy class with which the robot learns to include hierarchical components which can be extracted after learning and reused.  
This approach naturally fits recent research using deep neural networks for policy learning \cite{Nachum18,Vezhnevets17,Levy19}, which are sometimes structured to be successively executable in a manner similar to skill chaining \cite{Kumar18}, but have been used in other
more compact representations \cite{Daniel13,Daniel16}. Another approach aims to incrementally discover hierarchical skills that progressively learn to control factorized aspects of the environment, such as objects \cite{chuck2019hypothesis}.
These new results are exciting but have only just 
begun to scratch the surface of how component motor skills can be learned during the robot's task learning process.

\subsection{Learning Decision-Making Abstractions}

A collection of abstract motor skills provides \textit{procedural} abstraction to the robot; it can abstract over the low-level details of what it must \textit{do} to solve the task. However, these motor skills also provide an opportunity for abstracting
over the input the robot uses to decide which skill to execute. Learning such abstract structures have two potential advantages: First, the new abstract input may make learning for new tasks easier (or even unnecessary by enabling generalized abstract policies that work across tasks) and support abstract, task-level planning. Second, the resulting abstract representations may be much easier for a non-expert to edit, update, and interpret.

Broadly speaking, the literature contains two types of learned abstractions for decision-making. The first type learns an abstract policy directly,  while the second learns an abstract state space that serves as the basis for either planning or much faster learning in new tasks.

\textbf{Learning Abstract Policy Representations:} Here the goal is to learn a
generalized abstract policy that solves a class of problems. Often this policy encodes a mapping from abstract quantities to learned motor skills. For example, the policy for opening a door might include first checking to see if it is locked and if so, executing the unlock skill; then grasping the handle, turning it, and opening the door. Here the motor skills subsume the low-level variations in task execution (e.g., turning differently shaped doorknobs) while the abstract quantities subsume differences in task logic (e.g., checking to see if the door is locked). Approaches here differ primarily by the representation of the policy itself, ranging from finite-state machines \cite{niekum2015learning} to associative skill memories \cite{Pastor2012} to hierarchical task networks \cite{Hayes16} to context-free grammars \cite{lioutikov2018inducing}. In some cases these result in a natural and intuitive form of incremental policy repair and generalization \cite{niekum2015learning,Hayes16}. 

\textbf{Learning Abstract State Spaces:} Alternatively, the robot could learn abstract representations of \textit{state}, which when combined with abstract actions result in a new, but hopefully much simpler, and typically discrete, MDP. The robot can then use that simpler MDP to either learn faster, or to learn a task model and then plan. 
 Several approaches have been applied here, for example using the status of the skills available to the robot as a state space \cite{Brock05,Hart08}, finding representations that minimize planning loss \cite{Kulick2013,Jetchev13,Ugur2015,Ugur2015b}, using qualitative state abstractions \cite{Mugan12} and constructing compact MDPs that are provably sound and complete for task-level planning \cite{Konidaris16,konidaris2018skills,AmesTK18}. These approaches offer a natural means of abstracting away the low-level detail common to learning for manipulation tasks, and exploiting the structure common to task families to find minimal compact descriptions that support maximally efficient learning.

Taken together, hierarchical and compositional approaches have great promise for exploiting the structure in manipulation learning tasks, to reduce sample complexity and achieve generality, and have only begun to be carefully explored. This is a challenging  area full of important questions, and where several breakthroughs still remain to be made.

\section{Conclusion}
This paper has presented an overview of key manipulation challenges and the types of robot learning algorithms that have been developed to address these challenges. 
We explained how robots can represent objects and learn features hierarchically, and how these  features and object properties can be estimated using passive and interactive perception. 
We have discussed how the effects of manipulations can themselves be captured by learning continuous, discrete, and hybrid transition models. Different data collection strategies and model types determine how quickly the robot can learn the transition models and how well the models generalize to new scenarios. 

Skill policies can often be learned quickly from human demonstrations using behavioural cloning or apprenticeship learning, while mastery of manipulation skills often requires additional experience and can be achieved using model-based or model-free reinforcement learning. 
Given a skill, the robot can learn its preconditions and effects to capture its utility in different scenarios. Learning to monitor manipulation skills by detecting errors and goals imbues them with an additional level of robustness for working in unstructured environments. 

Finally, we have seen how a robot can exploit  the modular nature of manipulation tasks, such that each learned skill can be incorporated into a larger hierarchy to perform more advanced tasks, promoting reusability and robustness of skills across different tasks. 

Given the multi-faceted nature of manipulation, researchers have found great utility in being able to draw from methods that span the breadth of machine learning. However, 
despite access to these excellent generic machine learning methods, the challenges of learning robust and versatile manipulation skills are still far from being resolved. Some---but by no means all---of these pressing challenges are:
\begin{itemize}
    \item Integrating learning into complete control systems
    \item Using learned components as they are being learned (\textit{in~situ} learning)
    \item Safe learning, and learning with guarantees
    \item Exploiting and integrating multiple sensory modalities, including human cues
    \item Better exploration strategies, possibly based on explicit hypotheses or causal reasoning
    \item Exploiting common-sense physical knowledge
    \item Better algorithms for transfer across substantially different families of tasks
    \item Drastically improving the sample complexity of policy learning algorithms, while avoiding having to empirically tune hyper-parameters
\end{itemize}

As the community tackles these and other challenges, we expect that the core themes that have emerged repeatedly in manipulation learning research---modularity and hierarchy, generalization across objects, and the need for autonomous discovery---will continue playing a key role in designing effective solutions.

\acks{The authors would like to thank Animesh Garg for all of his help and insightful discussions, and
Ben Abbatematteo, Daniel Brown, Caleb Chuck, Matt Corsaro, Yuchen Cui, Eren Derman, Wonjoon Goo, Prasoon Goyal, Reymundo Gutierrez, Ajinkya Jain, Timothy Lee, Qiao (Jacky) Liang,  Ruldolf Liutikov, Matt Mason, Akanksha Saran, Mohit Sharma, and Garrett Warnell for their feedback and comments on the draft. 

This work was supported in part by the National Science Foundation (IIS-1724157, IIS-1638107, IIS-1617639, IIS-1749204, IIS-1717569), the Office of Naval Research (N00014-18-2243, N00014-18-1-2775) and the Sony Corporation.}

\bibliography{ManipulationRev}

\end{document}